\documentclass[12pt]{article}

\usepackage{newtxtext,newtxmath}
\usepackage{booktabs}
\usepackage{graphicx}
\usepackage{bm}
\usepackage{subcaption}
\usepackage{hyperref}
\usepackage[letterpaper,margin=1in]{geometry}

\renewenvironment{abstract}
	{\quotation}
	{\endquotation}

\date{}

\makeatletter
\renewcommand{\fnum@figure}{\textbf{Figure \thefigure}}
\renewcommand{\fnum@table}{\textbf{Table \thetable}}
\makeatother

\usepackage{scicite}

\usepackage{url}

\def\scititle{
MonoRace: Winning Champion-Level Drone \\ Racing with Robust Monocular AI

}
\title{\bfseries \boldmath \scititle}

\author{
    Stavrow A. Bahnam$^{\dagger}$,
    Robin Ferede$^{\ast,\dagger}$,
    Till M. Blaha,
    Anton E. Lang, \and
    Erin Lucassen,
    Quentin Missinne,
    Aderik E.C. Verraest,\and
    Christophe De Wagter, 
    Guido C.H.E. de Croon \and
	\small Control \& Operation, Delft University of Technology, Delft \& 2629 HS, The Netherlands.\and
	\small$^\ast$Corresponding author. Email: R.Ferede@tudelft.nl\and
	\small$^\dagger$These authors contributed equally to this work.
}

\begin{document} 

\maketitle

\begin{abstract} \bfseries \boldmath

Autonomous drone racing represents a major frontier in robotics research. It requires an Artificial Intelligence (AI) that can run on board light-weight flying robots under tight resource and time constraints, while pushing the physical system to its limits. The state of the art in this area consists of a system with a stereo camera and an inertial measurement unit (IMU) that beat human drone racing champions in a controlled indoor environment. Here, we present MonoRace: an onboard drone racing approach that uses a monocular, rolling-shutter camera and IMU that generalizes to a competition environment without any external motion tracking system. The approach features robust state estimation that combines neural-network-based gate segmentation with a drone model. Moreover, it includes an offline optimization procedure that leverages the known geometry of gates to refine any state estimation parameter. This offline optimization is based purely on onboard flight data and is important for fine-tuning the vital external camera calibration parameters. Furthermore, the guidance and control are performed by a neural network that foregoes inner loop controllers by directly sending motor commands. This small network runs on the flight controller at 500\ Hz. The proposed approach won the 2025 Abu Dhabi Autonomous Drone Racing Competition (A2RL), outperforming all competing AI teams and three human world champion pilots in a direct knockout tournament. It set a new milestone in autonomous drone racing research, reaching speeds up to 100 km/h on the competition track and successfully coping with problems such as camera interference and IMU saturation. 
\end{abstract}

\subsection*{INTRODUCTION}
\noindent
Autonomous drone racing (ADR) is a proving ground for high-speed perception, state estimation, planning, and control \cite{iros_2017, adr_survey}. The high speeds attained in drone racing represent a major challenge for the algorithms empowering autonomous flight. At such speeds, complex aerodynamic effects become increasingly difficult to model \cite{sun2018identification}, undermining both conventional control approaches and the simulation environments used for learning-based methods. In addition, computer vision in autonomous drone racing is considerably hampered by the large inter-frame displacements of imaged objects, resulting in high optical flow and motion blur. Furthermore, autonomous drone racing rewards pushing the physical drone system to its limits, requiring close-to-optimal control while being constrained by the requirement of fast, onboard processing.

In recent years, autonomous drone racing has seen rapid advances in control, perception, and state estimation. Early control approaches relied on differential-flatness-based trajectory tracking \cite{NIEUWSTADT19962301, MinSnap, Faessler2018, tal2020accurate}, followed by nonlinear model predictive control (NMPC) for faster and more aggressive flight \cite{aerospace4020031, Bicego2020, explicitMPC, torrente2021data, MPCC, TimeOpimalReplanning, MPC_DFBP, AdaptiveNMPC}. Today, neural network controllers trained with deep reinforcement learning represent the state of the art \cite{OCvsRL, Autonomous_Drone_Racing_with_Deep_Reinforcement_Learning, penicka2022learning, kaufmann2022benchmark, zurich_champion_level}, increasingly replacing modular trajectory-tracking pipelines. Recent work further replaces even low-level controllers with neural networks \cite{ferede2024supervised, izzo2024optimality, ferede2024end, ferede2025one}. On the perception side, the best results to date are obtained by stereo Visual Inertial Odometry (VIO) with geometric gate detection, fused through EKF-based state estimation \cite{zurich_champion_level, bosello2025your}. Although learned monocular VIO \cite{xu2021cnn, xu2025cuahn, bahnam2025self}, visual-model-predictive localization \cite{li2020autonomous, alphapilot_win_2019} and end-to-end vision-based policies \cite{geles2024demonstrating, xing2024bootstrapping, romero2025dream, krinner2025accelerating} are emerging as promising alternatives, they have not yet matched this state-of-the-art performance.

These advancements in autonomous drone racing have been stimulated by international competitions. Initially, these were purely an academic affair and organized alongside the IEEE IROS conferences. The first series of ADR competitions ran alongside IROS from 2016 to 2019: top speeds grew from 0.6 m/s (Daejeon, 2016) \cite{iros_2016} to 0.7 m/s (Vancouver, 2017) \cite{iros_2017}, 2.0 m/s (Madrid, 2018), and ~2.5 m/s (Macau, 2019). In 2019, Lockheed Martin and the Drone Racing League raised the bar with the AlphaPilot/AIRR series—a four-event competition with standardized hardware and a \$1~M prize. TU Delft’s MAVLab won the championship with an average velocity of 6.8 m/s and a peak of 9.2 m/s \cite{alphapilot_win_2019}, an order-of-magnitude jump over the early IROS speeds, while UZH’s runner-up reached up to 8 m/s on the same platform \cite{alphapilot_2e_2019}. Even so, human FPV drone racing pilots remained faster on comparable tracks. 

In 2022, the `Swift' approach \cite{zurich_champion_level} enabled an autonomous drone for the first time to beat champion-level pilots in head-to-head races. Perception and state estimation of Swift combined the Intel RealSense stereo Visual-Inertial Odometry (VIO), deep-neural-network-based gate detection on a front-looking camera, and a learned residual observation model trained with motion-capture supervision. Control was handled by a deep-RL policy that outputs the same low-level commands as human pilots (collective thrust and body rates). On an indoor track with peak speeds around 22 m/s \cite{adr_survey}, Swift won multiple heats and set the fastest recorded time, making it the first champion-level performance by an autonomous system. 

However, several challenges remain on the way to real-world application of this technology  \cite{zurich_champion_level,de2023drone}. Importantly, the AI for autonomous drone racing should not rely on external infrastructure such as a motion tracking system. Achieving independence from external infrastructure will reduce economic costs and substantially broaden the types of environments in which it can be applied. Moreover, it is desirable to further reduce the number of sensors used by the ADR solution, preferably only using a single forward-looking camera and onboard inertial sensors. 

The outstanding challenges in autonomous drone racing have led to the organization of a new international competition. The Abu Dhabi Autonomous Racing League (A2RL) and Drone Champions League (DCL) organized a new drone racing competition with a \$1 M price pool in April 2025. The goal of the competition was to push the state of the art by enforcing fully onboard perception with a single rolling-shutter CMOS camera and prohibiting any external aids at any stage. To this end, the organizers created the drone hardware accordingly and determined both the race track and the competition hall, which did not feature a motion tracking system. Nor did they allow any modification or the precision measurement of the track.

The competition was held on April 11 and 12, 2025, in Abu Dhabi. It comprised multiple events in which the autonomous racing drones competed with each other, with as main event being the ``Grand Challenge'' that focused on a single-drone time trial. At the same dates and in the same space, human FPV drone racing pilots competed in the DCL ``Falcon cup''. The best-ranked human FPV pilots were subsequently pitted against the best-ranked autonomous drones in a direct knockout tournament, starting with quarter finals (so in total three rounds till victory).

In this article, we present an onboard drone racing approach that only uses a single, rolling-shutter camera and an Inertial Measurement Unit (IMU). The approach won the above-mentioned Grand Challenge. Moreover, it enabled an autonomous drone to win for the first time an independently organized human vs. AI drone racing competition, sequentially beating three human FPV world champions in a direct knockout tournament. The approach includes a particularly robust state estimation pipeline that combines neural-network-based gate segmentation with a drone model. The robustness is illustrated by the fact that it can handle $50\%$ corrupted images. Moreover, the state estimation can cope with the saturation of accelerometers, which regularly occurred in the high-$g$ maneuvers during the very fast flight through the indoor racing track. Furthermore, we present an offline optimization procedure that leverages the known shapes and sizes of gates to refine any state estimation parameter. This offline optimization is based purely on onboard flight data and enables self-supervised post-flight fine-tuning of any vital parameters, like the external camera calibration.
The guidance and control in our system are executed by a single neural network, a Guidance-and-Control Network (G\&CNet), which maps the estimated state directly to control actions. Originally developed for optimal real-time spacecraft landing \cite{genc_space}, this concept has recently been adapted to high-speed quadrotor flight \cite{ferede2025one}. The G\&CNet is trained in simulation using reinforcement learning and directly outputs motor commands, fully replacing traditional inner-loop controllers. Until now, however, such nets relied on state estimates from highly accurate external motion-capture systems. In contrast, our approach uses only onboard vision-based state estimation. The resulting small network (only 3×64 neurons) runs on the 32-bit flight controller at 500 Hz and is able to successfully cross the reality gap. The proposed drone racing approach set a new milestone in autonomous drone racing research, beating all other AI systems and three human world-champion FPV racers, while reaching speeds up to 100 km/h on the competition track.


\begin{figure}[hbtp]
  \raggedright

    \begin{minipage}[b][6.1cm][t]{0.02\textwidth}
        \textbf{A}
    \end{minipage}%
    \begin{minipage}[b]{0.549\textwidth}
        \includegraphics[width=0.98\textwidth]{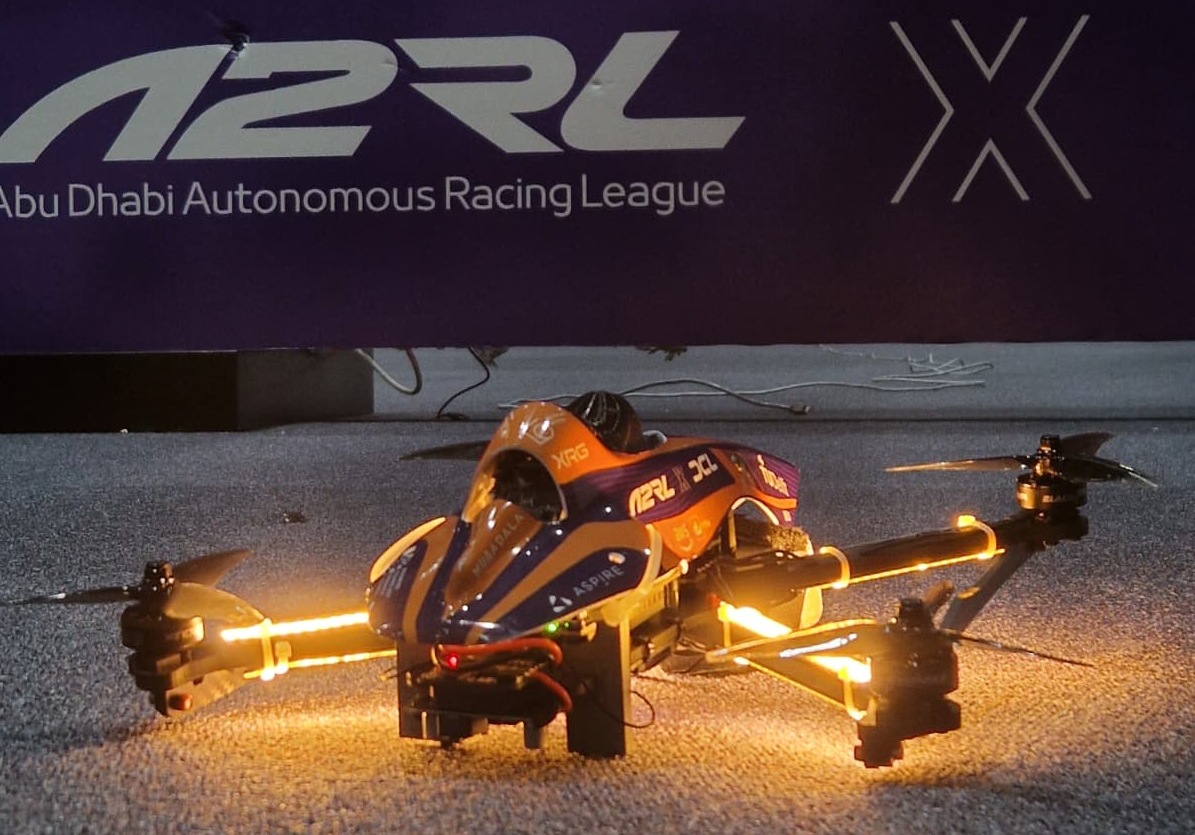}
    \end{minipage}%
    \begin{minipage}[b][6.1cm][t]{0.02\textwidth}
        \textbf{B}
    \end{minipage}%
    \begin{minipage}[b]{0.411\textwidth}
        \includegraphics[width=\textwidth]{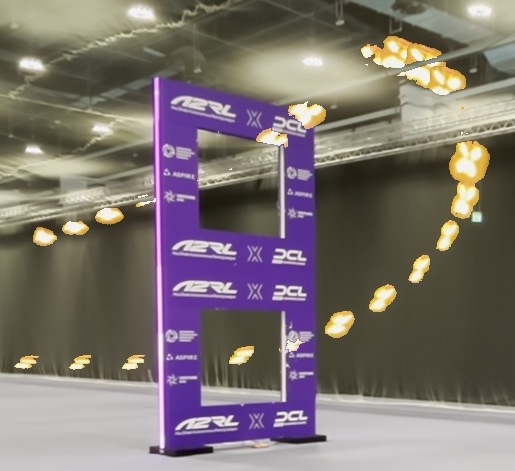}
    \end{minipage}

\vspace{3mm}
  \begin{minipage}[b][4.9cm][t]{0.02\textwidth}
    \textbf{C}
  \end{minipage}%
  \begin{minipage}[b]{0.98\textwidth}
    \includegraphics[width=\textwidth,trim={0 60 0 60},clip]{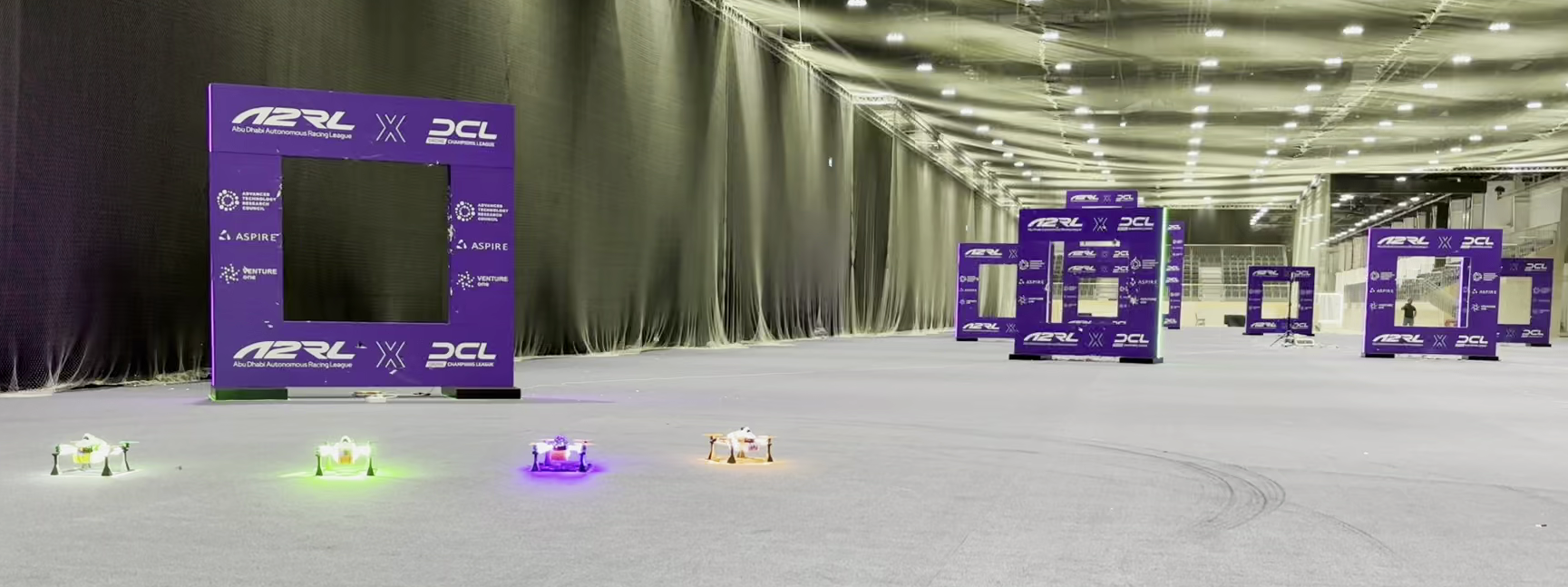}
  \end{minipage}

\vspace{3mm}

  \begin{minipage}[b][5.0cm][t]{0.02\textwidth}
    \textbf{D}
  \end{minipage}%
  \begin{minipage}[t]{0.96\textwidth}
    \includegraphics[width=\textwidth]{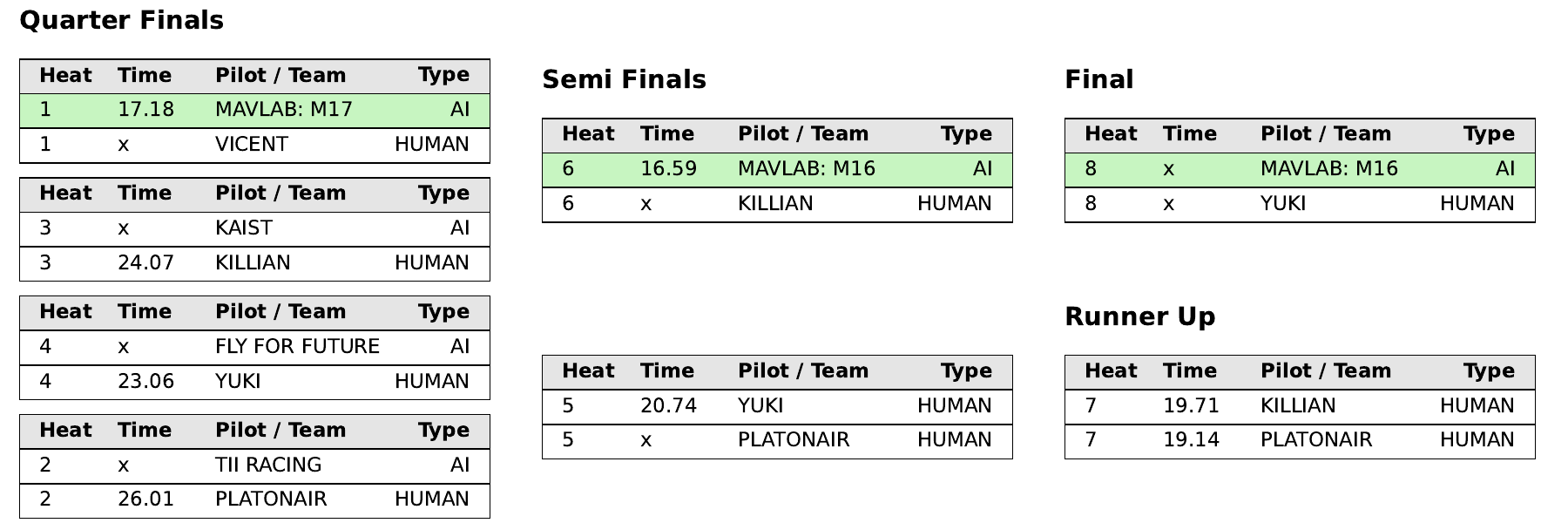}
  \end{minipage}

    \caption{(\textbf{A}) Close-up of the supplied externally made competition robot with its 5.1-inch propellers, its drone-race autopilot board with normal IMU, a single CMOS rolling-shutter camera connected to a GPU-equipped NVIDIA ORIN NX companion computer, and 8 bright orange LED strips to provide good visibility for the spectators. The robot is looking at an `A2RL x DCL' competition gate. (\textbf{B}) A time-lapse of the robot performing the split-S maneuver. (\textbf{C}) Overview of the entire track. The races consisted of two laps through eleven gates. Four robots are shown at the four start positions during a multi-robot race. (\textbf{D}) Results of the AI against human races.}
    \label{fig:Fig1}
\end{figure}

\subsection*{RESULTS}
\noindent
\subsubsection*{Competition Format}
The `A2RL x DCL' race was held in April 2025 in a 100 × 30 m indoor hall and featured a track with 11 square gates, each with a 1.5 m inner opening. A photograph of the track can be seen in Fig.~\ref{fig:Fig1}, included two double-gates and a split-S maneuver, where the drone passed through an upper gate before diving directly through a lower gate beneath it in the opposite direction. The objective was to complete two laps in minimal time. The organizers defined the completion time as the interval from the first passage of Gate 1 to the second passage of Gate 11. No external localization system was provided nor allowed, so all state estimation, control, and any fine-tuning had to rely solely on onboard sensing.

The event comprised the following four challenges:
\begin{enumerate}
    \item Grand Challenge — 15 minutes individual trial for AI teams (fastest time counts).
    \item AI vs Human — three knockout rounds in a head-to-head race (AI teams and FPV pilots).
    \item Multi-Drone Race — single race with four AI drones flying simultaneously.
    \item Drag Race — an 83\,m straight track with four gates, with a choice between two middle gates.
\end{enumerate}

Our system won the Grand Challenge, AI vs Human, and Drag Race, and placed third in the Multi-Drone Race due to the absence of collision-avoidance capabilities, while more than doubling the speed of the slowest drones. This paper focuses on the first two events. MonoRace is specifically designed for these scenarios, emphasizing high-speed, agile autonomous flight with onboard perception.

\subsubsection*{Performance}
\begin{figure}[hbtp]

   \begin{minipage}[b][6.0cm][t]{0.02\textwidth}
    \textbf{A}
  \end{minipage}%
  \begin{minipage}[t]{0.98\textwidth}
    \includegraphics[width=\textwidth]{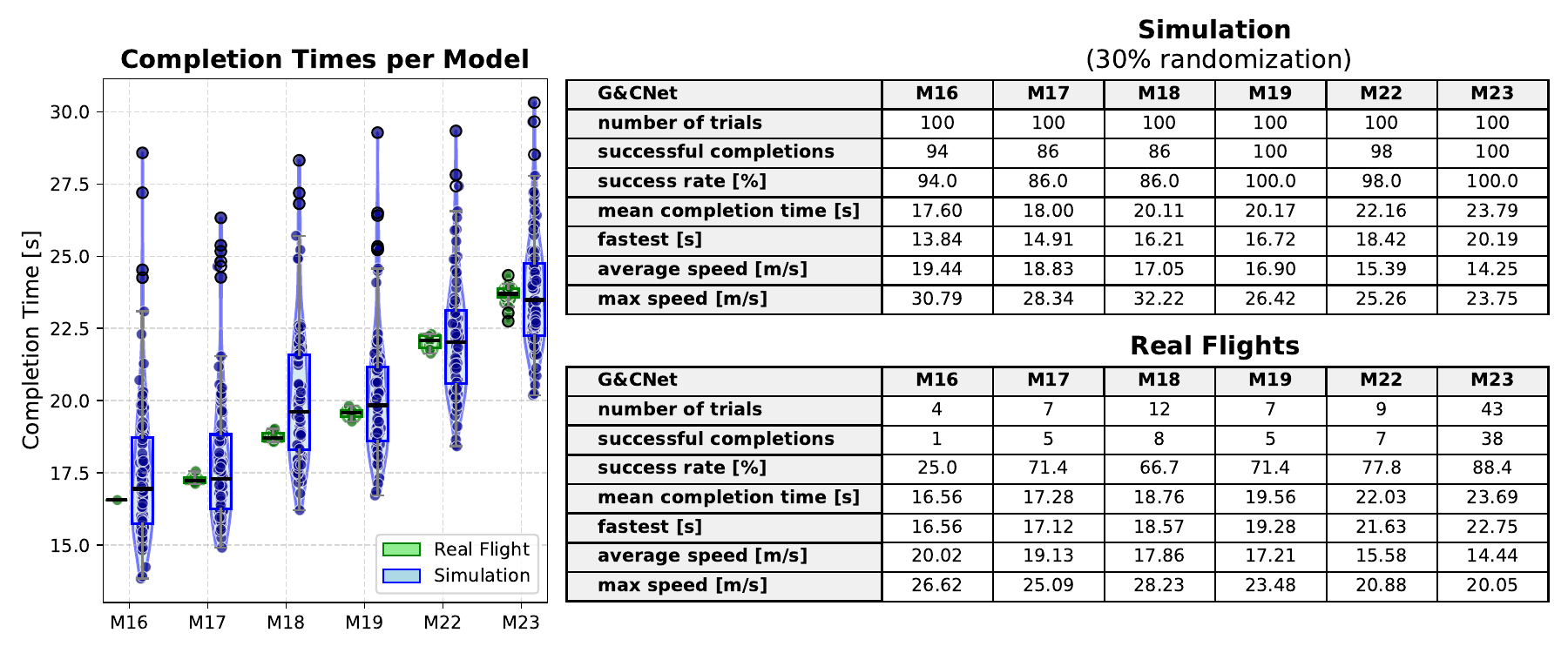}
  \end{minipage}

   \begin{minipage}[b][13.0cm][t]{0.02\textwidth}
    \textbf{B}
  \end{minipage}%
  \begin{minipage}[t]{0.98\textwidth}
    \includegraphics[width=\textwidth]{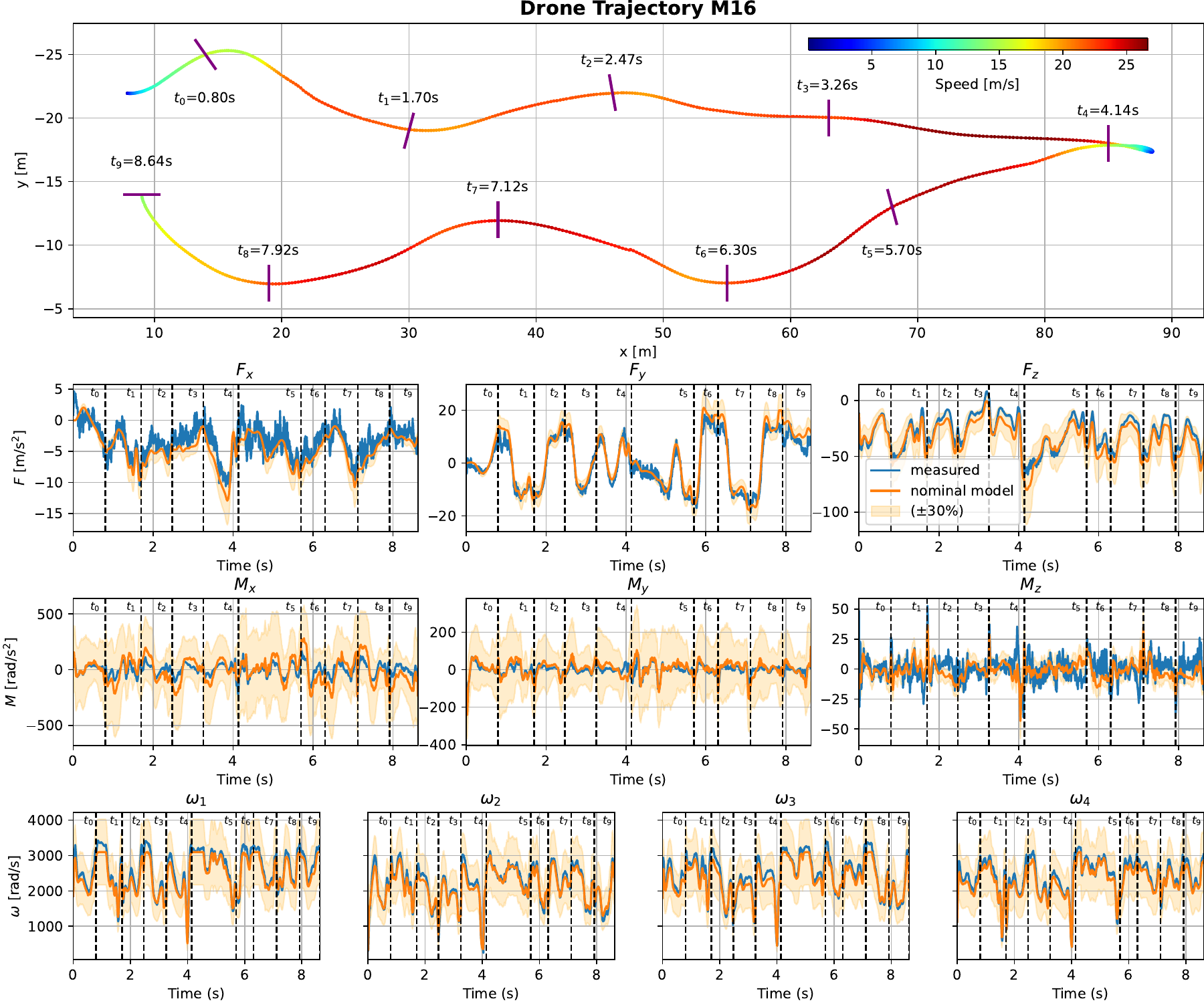}
  \end{minipage}

  \caption{(\textbf{A}) G\&CNet completion times: real flights versus simulation. (\textbf{B}) Fastest lap trajectory top view (M16) and the corresponding measured forces, moments, and actuator outputs versus the nominal model.}
  \label{fig:performance+reality_gap}
\end{figure}

Our controller achieved the fastest completion time of 16.56 s, outperforming three world champion-level FPV pilots.
The official organizer-recorded times\footnote{The times in Fig.~\ref{fig:Fig1} are provided by the race organizers and are measured using a VTx-based RF timing system~\cite{livetimescoring}. All other times reported in this paper use our onboard state-estimation–based timing, which explains the small numerical differences.} for the knockout rounds are shown in Fig.~\ref{fig:Fig1}. This result demonstrates that G\&CNets with low-level motor control can exceed human performance in high-speed drone racing. During the competition, we trained a variety of neural network controllers, summarized in Table~\ref{tab:policy_overview}. Each network was named after its fastest real-world completion time, rounded down to the nearest second and prefixed with an "M" (e.g., M16 for a policy completing the track in approximately 16 s). The comparison of lap times across different G\&CNets in Fig.~\ref{fig:performance+reality_gap} illustrates a clear trade-off between speed and robustness: more aggressive controllers consistently produced faster laps, whereas more conservative controllers prioritized safety and reliability. Remarkably, our most reliable network M23, completed 43 flights with an 88.4\% success rate, where most mishaps were development or hardware-related.


\renewcommand{\arraystretch}{0.7}
\begin{table}
\centering
\resizebox{0.7\textwidth}{!}{
\begin{tabular}{lccccccc}
\toprule
\textbf{Model} & &\textbf{M16} & \textbf{M17} & \textbf{M18} & \textbf{M19} & \textbf{M22} & \textbf{M23} \\
\midrule
Retrained from & & M17 & X & X & X & M23 & X \\
Initialization & & from ground & uniform & uniform & uniform & uniform & uniform \\
$\theta_{\text{cam}}$ (deg) & & 50 & 43 & 43 & 43 & 43 & 45 \\
$g_{\mathrm{size}}$ (m) & & $0.55^*$ & 0.45 & 0.50 & 0.45 & 0.40 & 0.40 \\
$g_{\mathrm{thickness}}$ (m) & & $0.8^*$ & 1 & 1 & 1 & 1 & 1 \\
$h_{\text{ground}}$ (m) & & 0.4 & 0 & 0 & 0 & 0 & 0 \\
$v_{\text{ground}}$ (m/s) & & 10 & 2 & 2 & 2 & 2 & 2 \\
\midrule
\multicolumn{7}{l}{\textbf{Rewards}} \\
$\lambda_{\text{prog}}$ & & 0 & 1 & 1 & 1 & 1 & 1 \\
$v_{max}$ & & - & 99 & 99 & 99 & 10 & 10 \\
$\lambda_{\text{gate}}$ & & 30 & 10 & 1 & 1 & 0 & 1.5 \\
$\lambda_{\Omega}$ & & 0 & 0.001 & 0.001 & 0.001 & 0.001 & 0.001 \\
$\lambda_{\text{offset}}$ & & 2 & 0 & 1 & 1 & 0 & 1.5 \\
$\lambda_{\text{perc}}$ & & 0.1 & 0.05 & 0.01 & 0.01 & 0.01 & 0.01 \\
$\lambda_{\Delta \bm{u}}$ & & 0.100 & 0.005 & 0 & 0 & 0 & 0 \\
$\Delta u_{\text{thresh}}$ & & 0.5 & 0.3 & 0 & 0 & 0 & 0 \\
$\lambda_{\bm{u}}$ & & 0 & 0 & 0 & 0.01 & 0 & 0 \\
$\lambda_{\text{crash}}$ & & 10 & 10 & 10 & 10 & 10 & 10 \\
\midrule
\textbf{Parameter} & \textbf{Nominal} & \multicolumn{6}{c}{\textbf{Randomization}} \\
$k_\omega$ & $1.55 \times 10^{-6}$ & 15 & 40 & 40 & 50 & 50 & 50 \\
$k_x$      & $5.37 \times 10^{-5}$ & 20 & 40 & 40 & 50 & 50 & 50 \\
$k_y$      & $5.37 \times 10^{-5}$ & 20 & 40 & 40 & 50 & 50 & 50 \\
$k_{x2}$   & $4.10 \times 10^{-3}$ & 45 & 40 & 40 & 50 & 50 & 50 \\
$k_{y2}$   & $1.51 \times 10^{-2}$ & 45 & 40 & 40 & 50 & 50 & 50 \\
$k_{\alpha}$ & $3.145$             & 45 & 40 & 40 & 50 & 50 & 50 \\
$k_{\text{hor}}$ & $7.245$         & 45 & 50 & 40 & 50 & 50 & 50 \\
$k_{p1}$   & $4.99 \times 10^{-5}$ & 45 & 40 & 40 & 50 & 50 & 50 \\
$k_{p2}$   & $3.78 \times 10^{-5}$ & 45 & 40 & 40 & 50 & 50 & 50 \\
$k_{p3}$   & $4.82 \times 10^{-5}$ & 45 & 40 & 40 & 50 & 50 & 50 \\
$k_{p4}$   & $3.83 \times 10^{-5}$ & 45 & 40 & 40 & 50 & 50 & 50 \\
$J_x$      & $-0.89$               & 50 & 40 & 40 & 50 & 50 & 50 \\
$k_{q1}$   & $2.05 \times 10^{-5}$ & 45 & 40 & 40 & 50 & 50 & 50 \\
$k_{q2}$   & $2.46 \times 10^{-5}$ & 45 & 40 & 40 & 50 & 50 & 50 \\
$k_{q3}$   & $2.02 \times 10^{-5}$ & 45 & 40 & 40 & 50 & 50 & 50 \\
$k_{q4}$   & $2.57 \times 10^{-5}$ & 45 & 40 & 40 & 50 & 50 & 50 \\
$J_y$      & $0.96$                & 50 & 50 & 40 & 50 & 50 & 50 \\
$k_{r1}$   & $3.38 \times 10^{-3}$ & 45 & 40 & 40 & 50 & 50 & 50 \\
$k_{r2}$   & $3.38 \times 10^{-3}$ & -  & 40 & 40 & 50 & 50 & 50 \\
$k_{r3}$   & $3.38 \times 10^{-3}$ & -  & 40 & 40 & 50 & 50 & 50 \\
$k_{r4}$   & $3.38 \times 10^{-3}$ & -  & 40 & 40 & 50 & 50 & 50 \\
$k_{r5}$   & $3.24 \times 10^{-4}$ & 45 & 40 & 40 & 50 & 50 & 50 \\
$k_{r6}$   & $3.24 \times 10^{-4}$ & -  & 40 & 40 & 50 & 50 & 50 \\
$k_{r7}$   & $3.24 \times 10^{-4}$ & -  & 40 & 40 & 50 & 50 & 50 \\
$k_{r8}$   & $3.24 \times 10^{-4}$ & -  & 40 & 40 & 50 & 50 & 50 \\
$J_z$      & $-0.34$               & 50 & 50 & 40 & 50 & 50 & 50 \\
$\omega_{\min}$ & $341.75$         & 20 & 40 & 40 & 50 & 50 & 50 \\
$\omega_{\max}$ & $3100.00$        & 15 & 33 & 30 & 30 & 30 & 40 \\
$k$         & $0.50$               & 45 & 50 & 40 & 50 & 50 & 50 \\
$\tau$      & $0.025$              & 30 & 50 & 40 & 50 & 50 & 55 \\
\midrule
\multicolumn{7}{l}{\textbf{PPO config}} \\
activation & & tanh & tanh & relu & tanh & relu & relu \\
layers $\pi$ & & 3 & 3 & 3 & 3 & 3 & 3 \\
neurons $\pi$ & &  64 & 64 & 64 & 64 & 64 & 64 \\
layers $V$ & & 3 & 3 & 3 & 3 & 3 & 3 \\
neurons $V$ & & 256 & 256 & 64 & 256 & 64 & 64 \\
ent\_coef & & 0.003 & 0.005 & 0.003 & 0.003 & 0 & 0 \\
\bottomrule
\end{tabular}
}
\caption{Training parameters for all models used in the competition.}
\label{tab:policy_overview}
\end{table}

Fig.~\ref{fig:performance+reality_gap} also presents the performance of 100 rollouts in our simulation environment for each network. Following the approach of \cite{ferede2025one}, we applied domain randomization to all model parameters; for this evaluation, we used 30\% randomization for all parameters. All networks maintained an approximately 90\% success rate despite substantial randomization. Robustness of the control for variations in drone properties is obtained by using even stronger randomization during training\footnote{LAP16 is an exception, as some parameters use less than 30\% randomization. However, most parameters are randomized $\ge$40\%, and evaluation with 30\% randomization still achieves a 94\% success rate.} (See parameters in Table~\ref{tab:policy_overview}). The results show that real-world completion times closely match the average simulated times, highlighting randomized simulation as a powerful and reliable tool for predicting real-world performance. Simulated crash rates, in contrast, were less predictive, as the simulator does not accurately model vision-based state estimation errors that occasionally caused real-world crashes.


The robustness of our approach becomes most evident when analyzing the trajectory of our fastest lap as seen in Fig.~\ref{fig:performance+reality_gap}. Measured specific forces, moments, and motor speeds deviated substantially from the nominal model but largely stayed within — and at times exceeded — the 30\% randomization bounds. Even in instances where the measurements exceeded these limits, the controller maintained stable flight and completed the lap. This ability to operate reliably beyond the nominal design envelope highlights the power of excessive randomization in training, allowing the drone to handle unmodeled dynamics while still achieving human-beating performance.

\subsubsection*{Evaluation and optimization with gate reprojections}
One of the great challenges of the A2RL competition is the absence of any external motion tracking or positioning system. This prevents the use of any ``ground truth'' position, which is typically used for fitting the drone's parameters or simply debugging the state estimation pipeline.
We therefore combine the drone's onboard images with the known sizes and shapes of the racing gates to still allow for evaluation and optimization in the absence of any external ground truth system.

Our proposed evaluation and optimization method is rooted in ``prediction error minimization'', a central concept in both state estimation filters \cite{simon2006optimal} and theories on the functioning of human brains \cite{friston2012embodied}. Ignoring observation noise, a perfect state estimate will result in a perfect correspondence between the expected observation of the world with the actual observation of the world. This prediction error, which is termed ``innovation'' in the control systems literature, is what leads state corrections in Kalman filters. Whereas typically this innovation is only used for correcting the state, one can actually also use it to optimize any part of the involved models and parameters. Suppose that we have two sets of state estimation filter parameters, $\Theta_1$ and $\Theta_2$, capturing aspects such as a drone's dynamics and its sensor properties (e.g., noise variance). If $\Theta_1 = \Theta^*$, i.e., it corresponds to the true dynamics and sensor properties, then we expect to have a smaller accumulated filtering innovation than when filtering the state with $\Theta_2 \neq \Theta^*$. This enables, for instance, evolutionary algorithms to optimize parameters like the variances of the different sensor measurements \cite{li2019unsupervised}, by minimizing the innovation over a given set of flight data. 

Although we did investigate the use of the filter's accumulated absolute innovation, we converged on a related but more sensitive metric that better captures long-term prediction quality, namely the correspondence between the expected racing gate pixels and the actually observed racing gate pixels in the gate segmentation images. Specifically, we use the state estimate together with the flight plan (map of the drone racing gates) to re-project gate masks in the image, and compare them with the actual observed gate segmentation with the help of the Intersection over Union (IoU). We then take the average of the IoU over one or more flights. This method can be used to assess the impact of changing any parameter in the state estimation pipeline, ranging from various parameters used in the acceptance of corner measurements to drone dynamics parameters. A higher resulting IoU indicates an improvement in the resulting state estimation.

Besides the evaluation of code changes, we also used the IoU in an offline optimization process. Although it can, in principle, optimize any parameter, we mainly used the method for the estimation of the extrinsic camera calibration parameters. Due to the flexible TPU camera mount, the angular offsets of the camera to the drone's body would regularly change, for instance, after any impact, modification or repair of the drone. The calibration setup present at the competition did not allow for a highly precise extrinsic calibration. However, over an entire flight, the onboard gate segmentation images provided enough evidence for the fine-tuning of these vital parameters. Hence, after a crash or repair, we flew the drone through the track with one of our safer networks, and subsequently optimized the camera extrinsic angles with the help of this IoU method. We used Bayesian optimization \cite{snoek2012practical}, since this allowed good results with limited samples - obeying the strict timing of the drone racing competition.

\begin{figure}[hbtp]
    \centering
    \raggedright
   \begin{minipage}[t]{0.95\textwidth}
    \textbf{A}\hfill
  \end{minipage}    \includegraphics[width=\linewidth, clip,trim={0 290 0 0}]{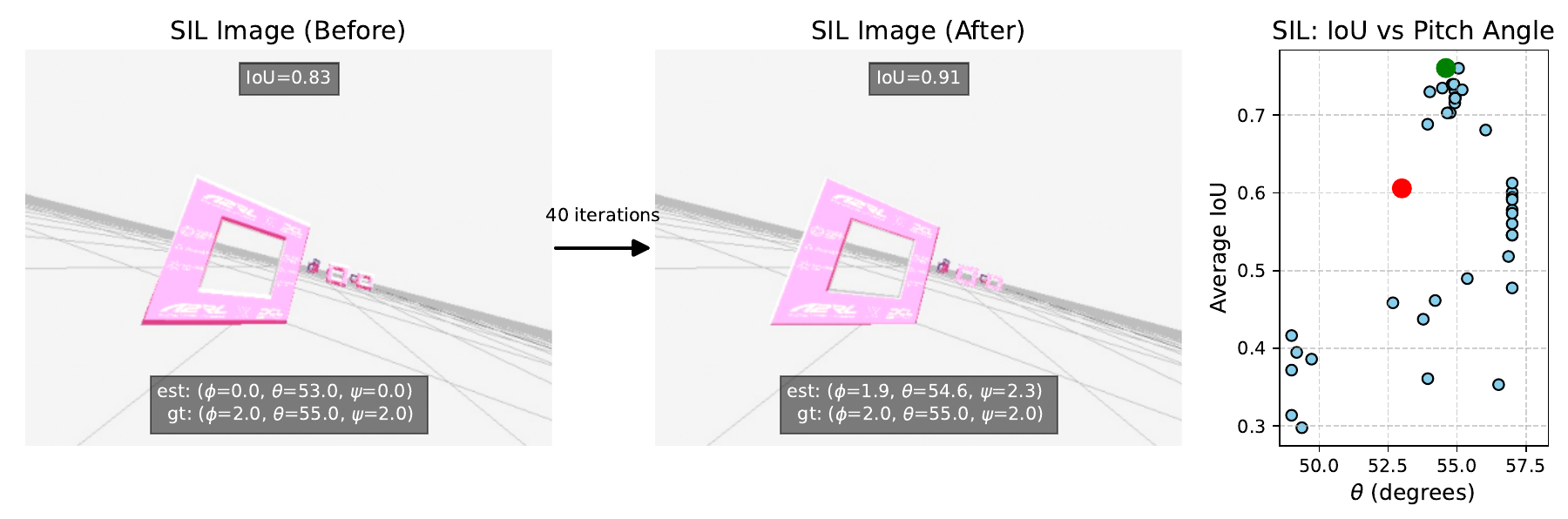}

    \raggedright
   \begin{minipage}[t]{0.95\textwidth}
    \textbf{B}\hfill
  \end{minipage}    \includegraphics[width=\linewidth, clip,trim={0 0 0 280}]{Figures/mask_optimization.pdf}
    
    \caption{Self-supervised refinement from onboard data only using mask-based extrinsic optimization. (\textbf{A}) Camera extrinsics optimization in the Software In the Loop (SIL) simulator. The estimated extrinsics align within $\approx 0.5^\circ$ of ground truth in only 40 steps optimizing a single log. (\textbf{B}) Result of the self-supervised refinement applied to camera extrinsics optimization on real flight data for only $40$ steps.
}
    \label{fig:mask_optimization}
\end{figure}

Fig.~\ref{fig:mask_optimization} shows the functioning of the mask-based camera extrinsics optimization method, applied both in simulation (top) and on real flight data (bottom). 
In the simulation experiment, we initialized the camera extrinsics with a 2-degree error relative to ground truth (for all angles). On the top left, the segmented gate (red) and the re-projected gate (white) do not fully overlap, yielding an average IoU of $0.83$. After optimization with $40$ iterations, the overlap improves significantly (average IoU $0.91$), and the estimated extrinsics $(1.9,54.6,2.3)$ are within $1^\circ$ of the ground truth $(2,55,2)$.


At the bottom of the figure, we applied the same method to real race data. Here, we take the extrinsics estimated by Kalibr\footnote{https://github.com/ethz-asl/kalibr} \cite{furgale2013unified} as comparison. The initial estimate $(-3.4,53.9,-4.5)$ deviates from the calibration $(-1.4,51.9,-2.5)$ (again 2 degrees offset per angle) and the overlap with the segmented gate is poor (average IoU $0.64$). After $40$ optimizer calls, the estimated extrinsics $(-0.8,51.9,-2.5)$ align much better with the Kalibr result, and the average IoU improves to $0.78$. This demonstrates that the optimization script can recover accurate extrinsics both in simulation and with real flight images, within about $1^\circ$ error. 
To test robustness, we ran this optimization for ten different single simulated flights with random $\mathcal{U}(-2^\circ, 2^\circ)$ perturbations in camera extrinsics.
The mean absolute errors of the optimized estimates were:
\[
\text{Pitch } 0.14^\circ\ (\sigma=0.09),\quad
\text{Roll } 0.49^\circ\ (\sigma=0.20),\quad
\text{Yaw } 0.71^\circ\ (\sigma=0.28)
\]
indicating sub-degree accuracy across runs.

\subsubsection*{IMU saturation}
The combination of an aggressive reinforcement learning (RL) controller capable of producing thrusts of up to $7g$, structural vibrations, and noise of the accelerometer led to reaching the accelerometer limit of $16g$. Upon saturation, the accelerometer provided measurements with sometimes extreme errors that quickly led to divergence in the IMU-based Kalman filter state, resulting in multiple crashes. This primarily occurred during the Split-S maneuver, but also after gate $1$ in the second lap, where we make a sharp turn with high velocity. To address this issue, we implemented a model-based acceleration prediction mechanism specifically activated during IMU saturation events. When the discrepancy between the measured accelerations and the drone's dynamic model-based estimates exceeds a predefined threshold, the Kalman filter incorporates these model-predicted accelerations instead of the saturated sensor readings. Additionally, we inflated the Kalman filter's uncertainty in both position and attitude states during such events to rely more on visual sensor measurements, thereby ensuring stable and accurate state estimation during aggressive flight maneuvers.

\begin{figure}[hbtp]
  \centering

\raggedright

 \begin{minipage}[b][4.8cm][t]{0.02\textwidth}
    \textbf{A}
  \end{minipage}%
  \begin{minipage}[t]{0.305\textwidth}
    \includegraphics[width=\textwidth]{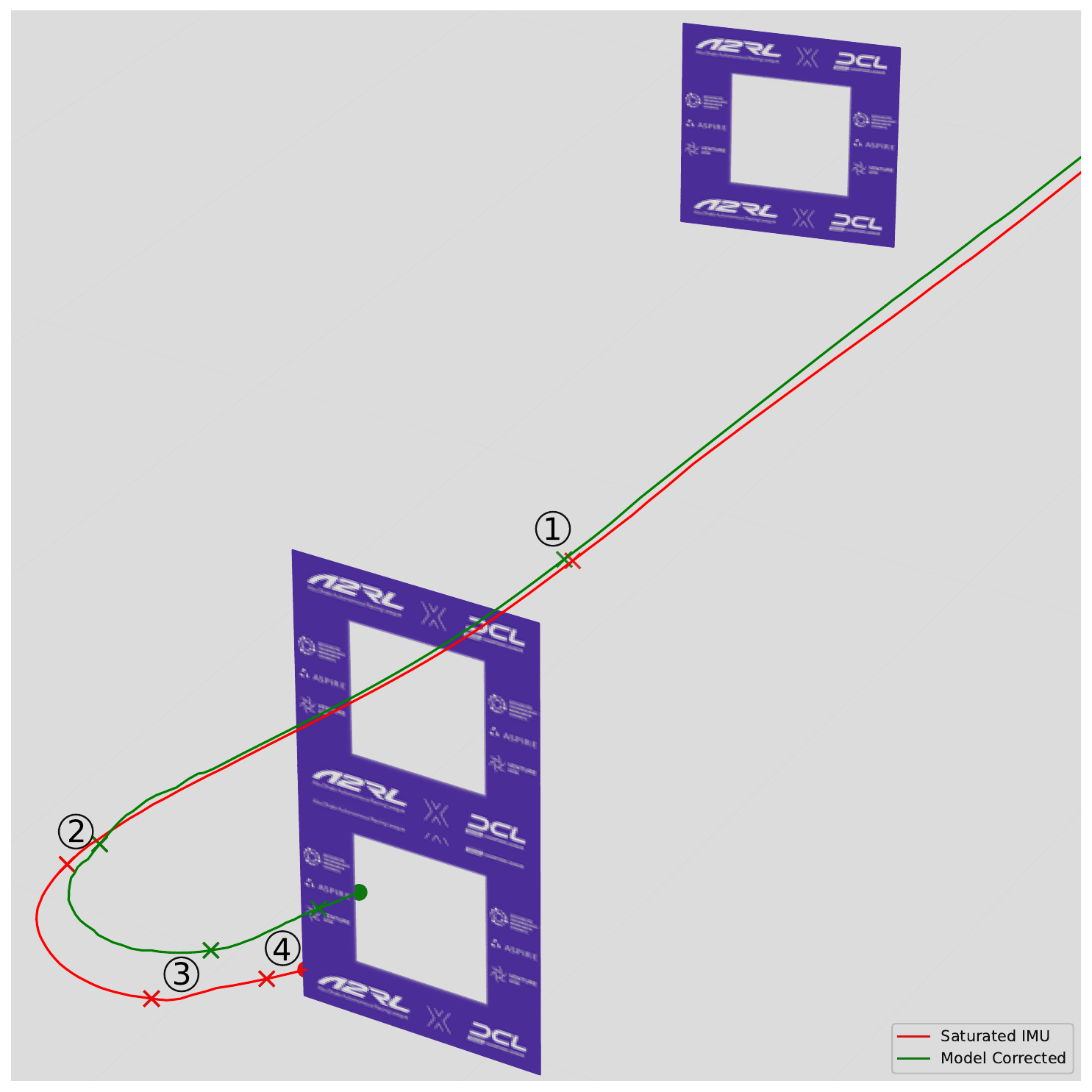}
  \end{minipage}\hfill
  \begin{minipage}[t]{0.675\textwidth}
    \includegraphics[width=\textwidth]{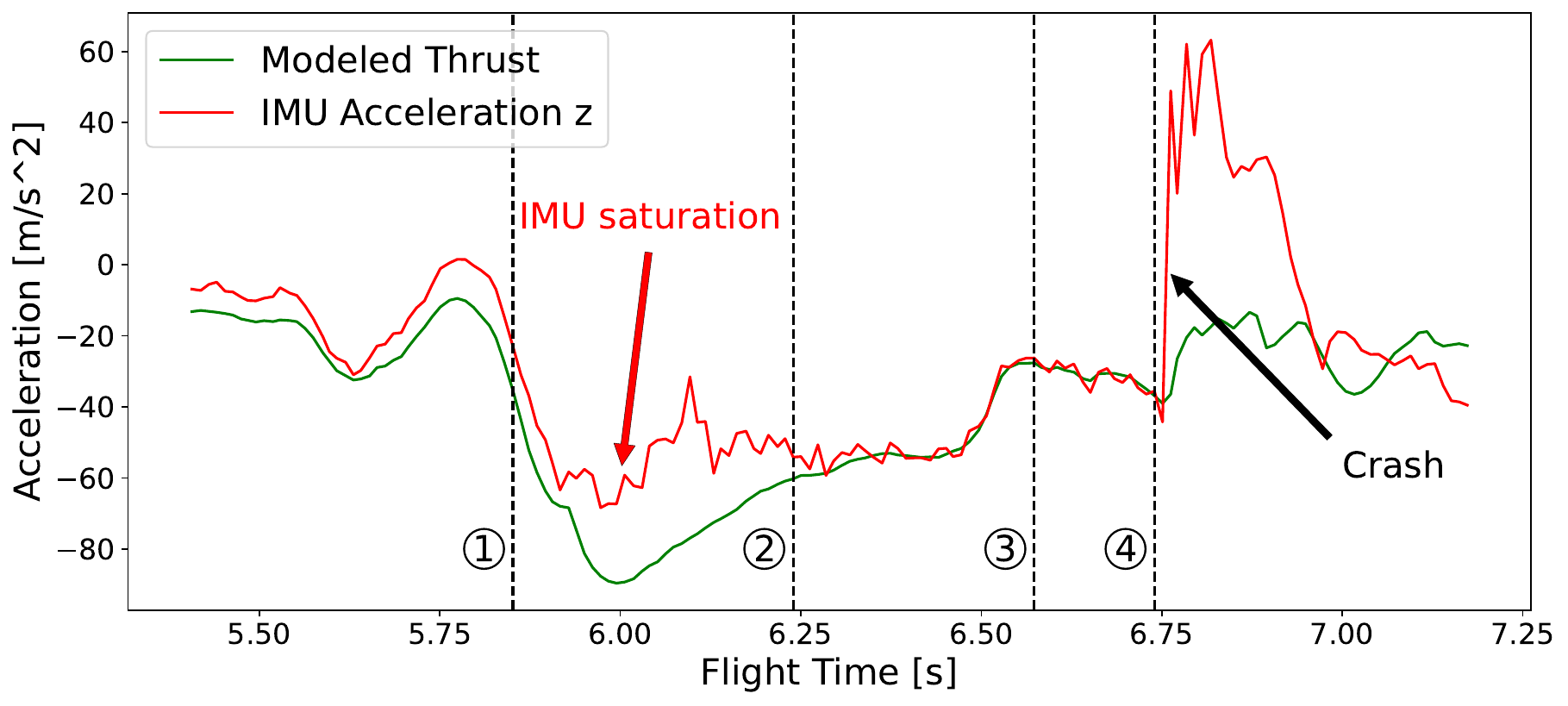}
  \end{minipage}

  \begin{minipage}[t]{\textwidth}
    \centering
    \includegraphics[width=1.0\textwidth]{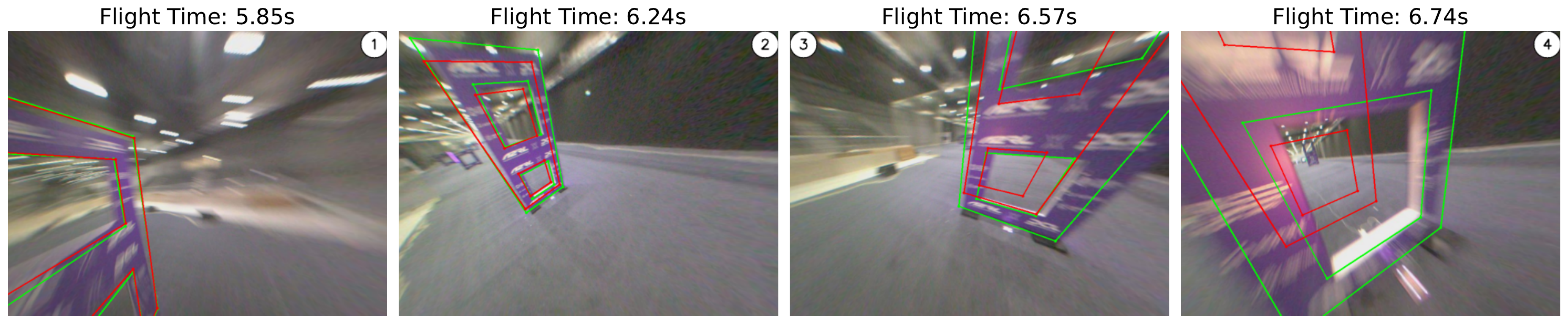}
  \end{minipage}

\raggedright

 \begin{minipage}[b][5.8cm][t]{0.02\textwidth}
    \textbf{B}
  \end{minipage}%
  \begin{minipage}[t]{0.425\textwidth}
    \includegraphics[width=\textwidth]{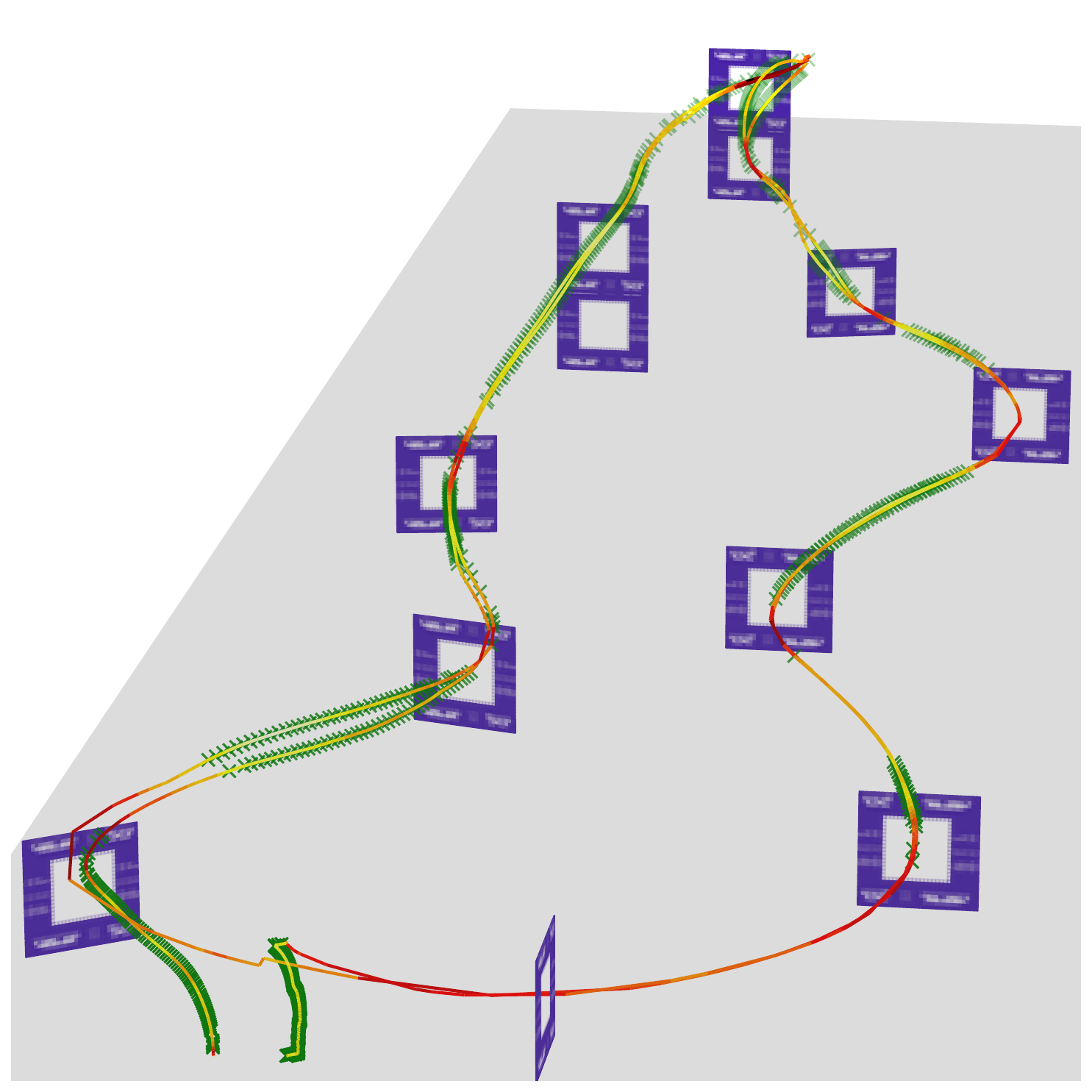}
  \end{minipage}\hfill
  \begin{minipage}[t]{0.555\textwidth}
    \includegraphics[width=\textwidth]{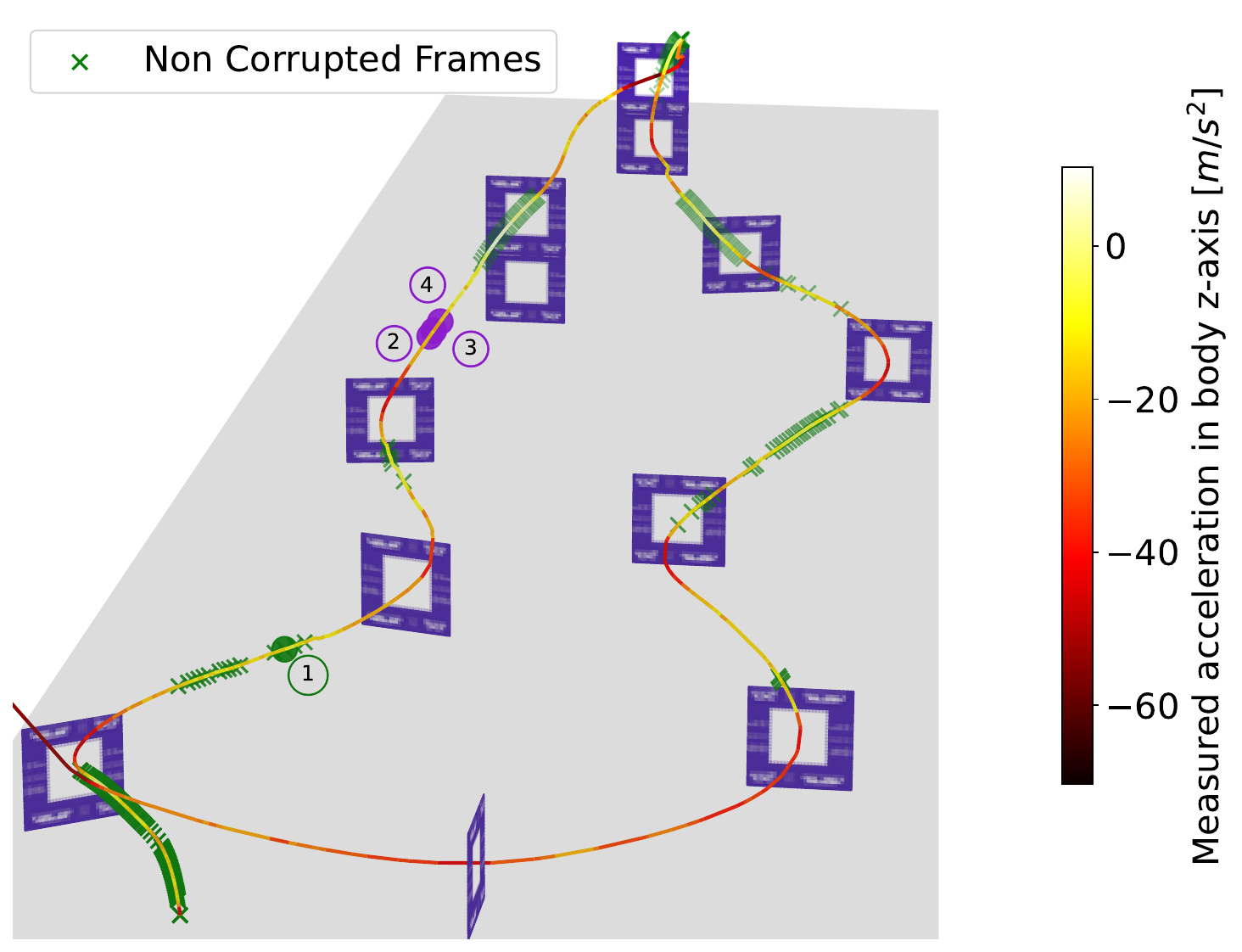}
  \end{minipage}
  
  \begin{minipage}[t]{\textwidth}
    \centering
    \includegraphics[width=1.0\textwidth]{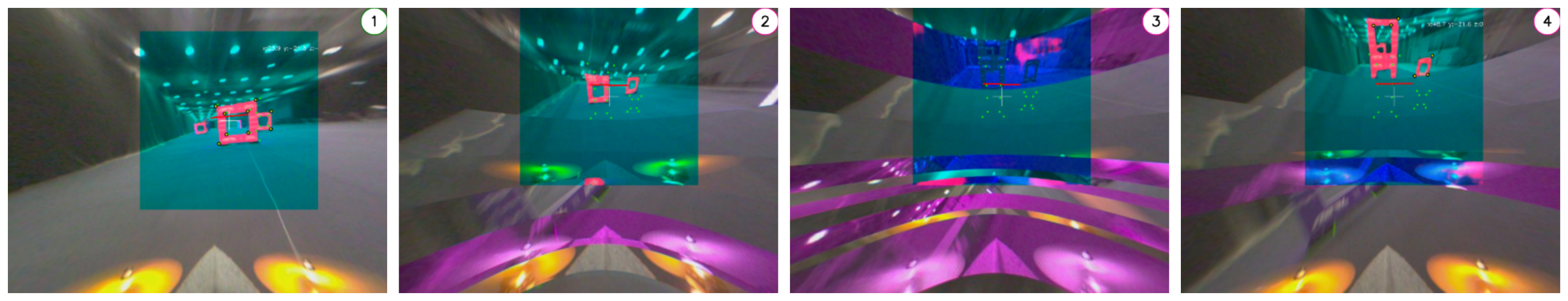}
  \end{minipage}

    \caption{Onboard sensor data corruption. (\textbf{A}) Comparison of the accelerometer-based versus model-based predicted trajectories, thrust and gate reprojections during a split-S with accelerometer saturation. (\textbf{B}) Successful dual lap completion despite 50\% image corruption (left) and single lap completion with 75\% image corruption from camera interference-caused purple horizontal bars, outdated top pixels and vertically shifted parts of the image.}

    \label{fig:corruption_trajectory}
\end{figure}

Fig.~\ref{fig:corruption_trajectory}a illustrates a trajectory segment where the drone crashed into the bottom gate of a Split-S maneuver due to IMU saturation in the body-z axis. The red trajectory represents state predictions derived from a Kalman filter (KF) using IMU data exclusively for the prediction step. The green trajectory shows state predictions from a Kalman filter that uses model-based acceleration estimates during IMU saturation. The model-corrected Kalman filter accurately predicts the trajectory of the drone, clearly indicating the imminent crash into the gate.

The adjacent acceleration plot compares the modeled thrust (green) and the measured IMU acceleration in the z-direction (red). Between timestamps $t_1$ and $t_2$, significant discrepancies are evident due to IMU saturation.

The bottom row shows gate reprojections: green indicates estimates from the model-corrected KF, and red indicates estimates from the IMU-predicted KF. Accurate state estimation is confirmed when the re-projected gates align closely with the actual gate positions in the images (purple). This visualization demonstrates that incorporating model-based predictions significantly improves the state estimate during aggressive maneuvers, such as the Split-S.

The controller that secured victory in the AI Grand Challenge, M17, achieved an overall success rate of $5$ out of $7$ flights. The two crashes were caused by IMU saturation. In fact, the IMU-only prediction resulted in $2$ successful flights and $2$ crashes, corresponding to a $50\%$ success rate. In contrast, incorporating model-based correction during IMU saturation led to a success rate of $100\%$, with all remaining 3 out of 3 flights successfully completed, on top of correctly measuring the crash in the previously failed logs.

\subsubsection*{Camera interference}
The drone designed by the organizers originally featured long MIPI cables for the camera connection. These cables were prone to bending under high thrust loads and occasionally came into contact with other hardware components, like the high noise SSD, resulting in electromagnetic interference. As illustrated in Fig.~\ref{fig:corruption_trajectory}b, this led to substantial image corruption that primarily occurred during periods of high acceleration in the z-axis. Across $10$ recorded flights, between $8\%$ and $75\%$ of captured images were corrupted. Two flights ended in crashes: one due to complete camera failure mid-flight, and another following a flight in which $75\%$ of the images were corrupted. In the latter case, the drone flew more than 25 m without valid images, successfully passing through one gate but crashing into a second gate, as shown in the right image of Fig.~\ref{fig:corruption_trajectory}b.

Although the hardware problem was later solved, 
MonoRace can address this type of problem thanks to its incorporated fail-safe strategies. First, the gate segmentation network does not find gates in the corrupted parts of the frames (Fig.~\ref{fig:corruption_trajectory}b.3). Next, a RANSAC-based outlier rejection matches the 2D affine transformation between priors to corner candidates to reject erroneous corners (Fig.~\ref{fig:corruption_trajectory}b.2). Finally, our KF also filters out measurements that deviate too much from the predicted state, based on the uncertainty level of the Kalman Filter (Fig.~\ref{fig:corruption_trajectory}b.4). Combined, this allowed the drone to survive most of the interference, while still refining the position with any occasional good frame.

\subsection*{DISCUSSION}
\noindent

In this work, we have presented MonoRace: a fully onboard, monocular, rolling-shutter perception-control pipeline for autonomous drone racing that won the April 2025 Abu Dhabi Autonomous Drone Racing Competition (A2RL), outperforming all competing AI teams and three world champion FPV pilots in direct knockout heats. On the competition track, the vehicle reached a maximum speed of 28.23 m/s, which, to our knowledge, is the fastest fully onboard, autonomous flight reported to date. For context, \cite{adr_survey} reports 22 m/s on a different $30\times30\times8$ m track \cite{zurich_champion_level}; our results were obtained on a
$76\times18\times5.4$ m track, using a monocular CMOS camera, a saturating IMU and no external aids at any stage.

A key element is the Guidance-and-Control Network (G\&CNet) that directly outputs motor commands, paired with vision-based state estimation. Extensive domain randomization helped bridge the simulation-reality gap at high speeds where aerodynamic effects are less predictable \cite{Loquercio2020domainrandom}. By bypassing intermediate control loops, the G\&CNet achieved a very low end-to-end latency; the controller operated at 500 Hz with full throttle authority, allowing the motor commands to change from 0-100$\%$ in $2$ ms.

The monocular visual state estimator proved robust to camera interference and IMU saturation. The drone successfully completed the full track with $50\%$ of the frames being corrupted. During periods of high
$g$ loading and structural vibration that caused IMU saturation, we replaced measured accelerations with model predictions, which stabilized the estimator.

Although the proposed drone racing approach set a new milestone, several improvements are still possible. First, the vision and control remain decoupled. On the one hand, this results in a larger necessity for reward shaping, e.g., including an explicit penalty for not looking at the gates. On the other hand, the control network expects perfect state estimation, while the estimated state is always uncertain to a given extent. A possible solution avenue is to perform end-to-end learning. Second, the vision pipeline heavily relies on the rectangular shape of the gates for both the corner detection and the localization relative to the gate. In contrast, humans can easily adapt to different gate shapes, with no training necessary. In order to deal with different gate shapes, one could train a neural network to directly map an image to relative gate poses. It would be straightforward to artificially generate a dataset for this end. However, refinement in the real world without any available ground-truth is less evident. The mask reprojection idea could perhaps be used to retrieve the ground truth relative positions needed for such refinement. Third, the proposed approach does not yet detect and avoid other drones on the track. This did not hamper its performance in the human vs. AI tournament, but it did result in a crash during the multi-AI event, in which four AI drones flew the track at the same time. In that event, our drone took off first, flying fast, but then had to double the other drones. Without drone detection and avoidance, it flew into the drone of another team, resulting in a 3$^{\mathrm{rd}}$ place in that event. The main challenge here will be to detect and avoid other drones, while not adding too much computational effort and complexity to the vision and control pipeline. 

Although the proposed approach focuses on agile flight for drone racing, it will have a much broader impact. The proposed guidance and control neural network is small enough to comfortably run on a microcontroller and still approximate time-optimal flight. The required computation time for this is a fraction of traditional pipelines that first plan trajectories and then attempt to execute them with the help of online guidance and control. The current study illustrates that G\&CNets have the potential to unlock optimal control for even the smallest and cheapest robots.

In this article, the application is FPV drone racing, an e-sports in which drones may play a similar role in the future as chess programs play in the board game of chess today: to assist human players in training and unlocking novel moves and strategies that have not yet been thought of by human beings. Further generalizing the approach away from drone-race-specific elements, such as racing gates, will allow for many more real-world applications. Of course, the extremely fast flight speeds immediately suggest military applications, where fast flight has not only offensive but also defensive utility. However, we hope and expect the main application to lie in the societal domain; Currently, fully autonomous drones are typically still rather slow, which means that they fly at speeds that are sub-optimal for energy usage. Slight changes to the reward functions can enable smooth and swift flight, with the ability to adapt speed to the requirements of the environment and the mission. This will, in general, allow drones to perform longer-range and duration missions, while slowing down when safety requires it.

\section*{MATERIALS AND METHODS}
\noindent

\subsection*{Overview}
\begin{figure}[hbtp]
    \centering
    \includegraphics[width=1.0\linewidth]{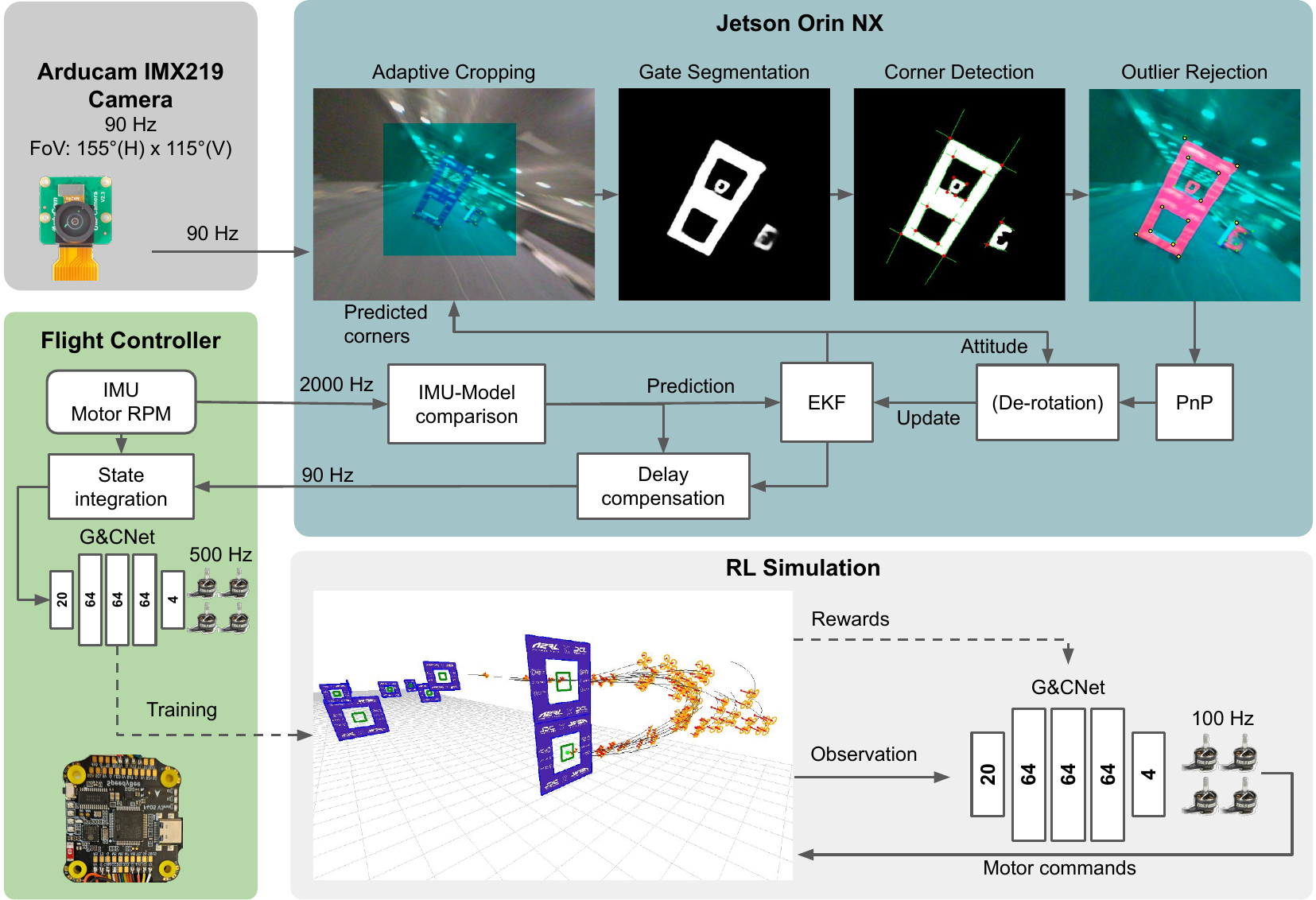}
    \caption{Overview of the pipeline and physical location of various elements. An IMX219 camera image is grabbed and timestamped on the Orin. Adaptive state-based cropping selects the part of the wide-field-of-view image with the best statistics for detecting corners. Gate segmentation is performed on the GPU, and subsequently, precise corner detection is performed from edge identification. After outlier rejection, the relative pose is computed with respect to the assumed gate location. After sensor fusion with delay compensation, the state is sent to the flight controller, which runs a 500Hz local state estimation filter and the direct-to-motor neural G\&CNet, which was trained in RL simulation with randomization and is being executed with zero-shot transfer.}
    \label{fig:overview}
\end{figure}

Here, we provide a brief overview of the proposed approach. A high-level summary of the method is illustrated in Fig.~\ref{fig:overview}. We introduce hardware, the state-estimation algorithm, and the reinforcement-learning-based control strategy. A more detailed explanation of each component is provided in the Perception and Control sections.


The drone uses a custom carbon fiber frame designed by the competition organizers and has a total mass of 966 g. Its sensor suite consists of a monocular rolling-shutter camera with a wide field of view (155° horizontal × 115° vertical), and an IMU integrated into the flight controller. The IMU provides accelerometer measurements at 1000 Hz and gyroscope measurements at 2000 Hz. All computation is performed onboard, using an NVIDIA Jetson Orin NX as the main processing unit.

Our state estimation approach begins by capturing images of $820\times616$ at 90\ Hz, which are adaptively cropped and resized to $384\times384$ based on the predicted gate locations. The resulting crops are processed by our gate-segmentation model, GateNet. From the resulting segmentation masks, our gate-specific corner detector, QuAdGate, extracts gate edges and computes their intersections to accurately localize gate corners. After corner detection, we apply homography-based outlier rejection between the detected corners and the expected corner projections derived from the prior state estimate. Finally, we estimate the drone’s pose using a Perspective-n-Point (PnP) solution combined with attitude estimates from the Extended Kalman Filter (EKF). The pose estimates from visual processing are then fused with high-rate IMU data in an EKF to ensure robust and accurate state estimation even during aggressive flight maneuvers. Due to the high $g$-loads and structural vibrations, the accelerometer saturates. To mitigate this, we employ a dynamic drone model to detect IMU saturation and correct the state predictions accordingly.

Finally, the estimated states are used as an input to a Guidance \& Control Network (G\&CNet), which operates at 500 Hz and directly outputs motor commands. This light neural network runs on the flight controller itself, which is equipped with an STM32H743 ARM processor (480 MHz). The network distinguishes itself through its real-time responsiveness and adaptability, enabling precise trajectory tracking, aggressive maneuvering and instant re-planning in complex, high-speed racing environments. It allows faster flight by removing the unpredictability of a traditional inner-loop controller near saturation \cite{ferede2024end}. The networks are trained with reinforcement learning in a simplified simulation of the race drone that captures the dominant actuator dynamics, aerodynamic forces, and moments. Using this simulation together with a reward function, the policy is optimized via PPO to complete the track in minimal time. To support robust sim-to-real transfer, broad domain randomization is applied to all model parameters, following the approach of \cite{ferede2025one}. During the competition, various combinations of randomization ranges, reward functions, and PPO hyperparameters were explored to achieve different speed/robustness trade-offs. The network with the fastest lap time was ultimately deployed against the human pilots, which it successfully outperformed.

\subsection*{Perception}

\subsubsection*{Adaptive Cropping}

We use a modified camera driver to capture images at 90 Hz with the full field of view of $155^\circ \times 115^\circ$ and a resolution of $820 \times 616$ pixels. As a first preprocessing step, the images are undistorted. To reduce computational load and enable real-time processing on the Jetson Orin NX at 90\ Hz, we subsequently limit our vision pipeline to a $384 \times 384$ pixel region ($29.2\%$ of the original image).

For each incoming frame, we predict the expected pixel locations of all gate corners using the current state estimate. Based on visibility and distance, we select the image area with the two closest visible gates for further processing. Gates are excluded when viewed at an excessively oblique angle that would result in strong aspect-ratio distortions; in such situations, even small pixel-level errors can induce large errors in the estimated lateral position and heading estimation. We also discard gates when the projected center of the gate lies outside the image boundaries.

If the predicted gate corners are within a $384 \times 384$ region of the full-resolution image, we directly crop that region. Otherwise, we first resize the image and then crop the $384 \times 384$ window around the predicted gate location. This adaptive cropping strategy improves computational efficiency while minimizing loss of pixel accuracy, ensuring that the perception module receives a consistent input resolution regardless of gate distance or viewing angle. It effectively analyzes distant gates with original resolution while downscaling large, close gates.

In Fig.~\ref{fig:vision}C,
we compare several strategies for feeding the original image to the $384 \times 384$ pixel network, analyzed on the three fastest successful flights in the competition. The first option is simply resizing the full image to $384 \times 384$ (`Resized') without preservation of the horizontal–vertical aspect ratio (AR). `Resized fixed AR' refers to first scaling the image to $511 \times 384$, followed by a $384 \times 384$ crop. `Center cropping' directly crops a $384 \times 384$ region from the center of the original image without resizing.
`Adaptive cropping' achieves the best performance across all metrics: it obtains the highest IoU with the gate segmentation, detects $36\%$ more corners than the simple resized approach, and yields $25\%$ more frames with sufficient points for a valid PnP solution. The adaptive cropping is able to detect more corners, especially in gates that are far away, as these become too small to detect when resizing the image.
In contrast, center cropping performs substantially worse due to the reduction in the field of view combined with the very high pitch angles. In one of the three evaluated flights, this method even led to divergence of the Kalman filter, which in turn explains the markedly lower IoU observed for the center-cropping approach.

\subsubsection*{GateNet}
\paragraph{Network}
For gate segmentation, we adopt a U-Net–style architecture \cite{unet_original}, inspired by prior work \cite{alphapilot_win_2019}, which we refer to as GateNet. This network consists of an encoder–decoder with skip connections from the encoder to the corresponding decoder layer, following the standard U-Net design, where channel dimensions are scaled by a factor $f$. Each block consists of double $3 \times 3$ convolutional layers with batch normalization and ReLU activation functions.

The network structure first consists of initial double convolutional blocks, denoted as \texttt{inc}. \texttt{downk} denotes a max-pooling layer followed by a double convolutional block with $k$ output channels. Before applying \texttt{downk}, the corresponding features are stored as skip connections for the decoder. \texttt{upk} denotes a transposed convolution and batch normalization, followed by adding the skip connections from the corresponding encoder layer, and then a double convolutional block with $k$ output channels. Finally, \texttt{outc-1} denotes a $1 \times 1$ convolution mapping to a single output channel and a sigmoid activation function. The resulting network is defined as:

\begin{align*}
\begin{array}{c c c}
\text{Encoder} & \text{Decoder} & \text{Outputs} \\[2mm]
\texttt{inc-64/f} & \rightarrow \texttt{up4-64/f} & \rightarrow \texttt{outc4-1} \\[1mm]
\downarrow & \uparrow & \\[1mm]
\texttt{down1-128/f} & \rightarrow \texttt{up3-64/f} & \rightarrow \texttt{outc3-1} \\[1mm]
\downarrow & \uparrow & \\[1mm]
\texttt{down2-256/f} & \rightarrow \texttt{up2-128/f} & \rightarrow \texttt{outc2-1} \\[1mm]
\downarrow & \uparrow & \\[1mm]
\texttt{down3-512/f} & \rightarrow \texttt{up1-256/f} & \rightarrow \texttt{outc1-1} \\[1mm]
\downarrow & \nearrow & \\[1mm]
\texttt{down4-512/f} & & \rightarrow \texttt{outc0-1} \\
\end{array}
\end{align*}

GateNet produces five output maps $\{y_0, y_1, y_2, y_3, y_4\}$ at progressively increasing resolutions. These multi-scale predictions are supervised using auxiliary losses between each output map and appropriately down-scaled ground-truth masks, improving gradient flow and overall segmentation accuracy. During deployment, only the highest-resolution output map is used.

\paragraph{Training setup}
For both training and deployment, we use a resolution of $384 \times 384$ and $f = 4$. Before training, Xavier uniform weight initialization \cite{xavier_init} is applied. GateNet is trained for $100$ epochs with a base learning rate of $1 \cdot 10^{-3}$, reduced by a factor of $\sqrt{0.1}$ at epochs $10$, $33$, $66$ and $90$, resulting in a final learning rate of $1 \cdot 10^{-5}$. Training is performed with the AdamW optimizer and a batch size of $16$. Each output map is supervised using a combination of Dice loss and binary cross-entropy (BCE) loss:  
\begin{align*}
\mathcal{L}_i = \mathcal{L}_{\text{Dice}}(y_i, \hat{y}_i) + 2 \cdot \mathcal{L}_{\text{BCE}}(y_i, \hat{y}_i) 
\end{align*}
Finally, we apply output-specific scaling factors to emphasize higher-resolution predictions:
\begin{align*}
\mathcal{L}_{\text{tot}} = 4 \cdot \mathcal{L}_0 + 2 \cdot \mathcal{L}_1 + \sum_{i=2}^{4} \mathcal{L}_i.
\end{align*}

\paragraph{Synthetic data}

To reduce the need for manual labeling and improve generalization across different track layouts, we generate synthetic data in addition to manually labeled real-world images. In our synthetic data pipeline, each gate face is subjected to a series of randomized transformations, including scaling, rotation, and perspective warping to simulate varied viewpoints and in-flight camera distortions. Once transformed, the gate is composited onto a background image sampled from a curated set of representative environments (warehouses, parking lots, empty halls, etc.)

To increase photometric variability and simulate lighting changes, both gate and background images undergo HSV-based color augmentations independently. This includes random shifts in hue, saturation, and value to mimic different lighting conditions and material reflectances. Furthermore, we introduce image-level noise by applying Gaussian noise with randomly sampled variance across images. This noise emulates sensor imperfections like thermal noise \cite{cmos_cam_noise_2008} together with environmental artifacts, and contributes to the robustness of the trained model. All augmentation strategies were applied in a randomized manner per sample, ensuring a high degree of variation across the synthetic dataset (See Fig.~\ref{fig:vision}B). The resulting images form a diverse and richly augmented set that helps bridge the domain gap between synthetic and real-world data, and operate the robot in new environments without any additional image labeling.

\paragraph{Data augmentations}

The synthetic and real-world datasets are merged and fed through a unified dataloader (in a 3500:500 synthetic to real split) to ensure consistent preprocessing and augmentation across all samples. During training, each image undergoes randomized affine transformations—comprising rotation, translation, and scaling—to enhance spatial variability and viewpoint diversity. Additional photometric augmentations are performed in the HSV color space to also vary labeled images, introducing random perturbations in hue, saturation, and value to emulate lighting variations and material appearance changes.

To further mimic the visual artifacts encountered during high-speed flight, artificial motion blur is applied by convolving each image with a square averaging kernel of random size (5–15 pixels) and random orientation. Complementary noise models, including thermal noise and kernel-based blurring, are also introduced to replicate sensor noise and lens-induced degradation. These augmentations, applied stochastically at load time, yield a more diverse and robust training distribution that facilitates improved generalization.

\subsubsection*{QuAdGate}

\begin{figure}[p]

\raggedright

 \begin{minipage}[b][1.5cm][t]{0.02\textwidth}
    \textbf{A}
  \end{minipage}
  \begin{minipage}[t]{0.95\textwidth}
    \includegraphics[width=\linewidth]{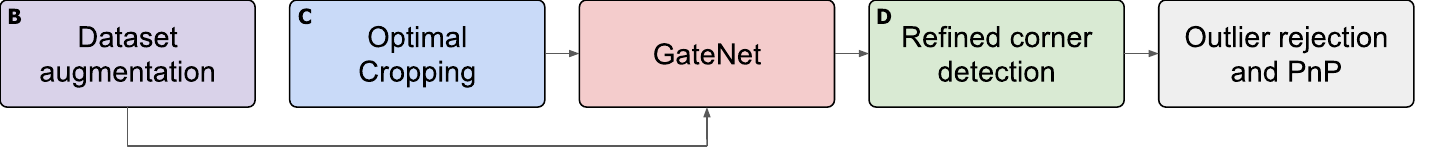}
  \end{minipage}

\raggedright

 \begin{minipage}[b][6.7cm][t]{0.02\textwidth}
    \textbf{B}
  \end{minipage}
  \begin{minipage}[t]{0.95\textwidth}
    \includegraphics[width=\linewidth, height=7cm]{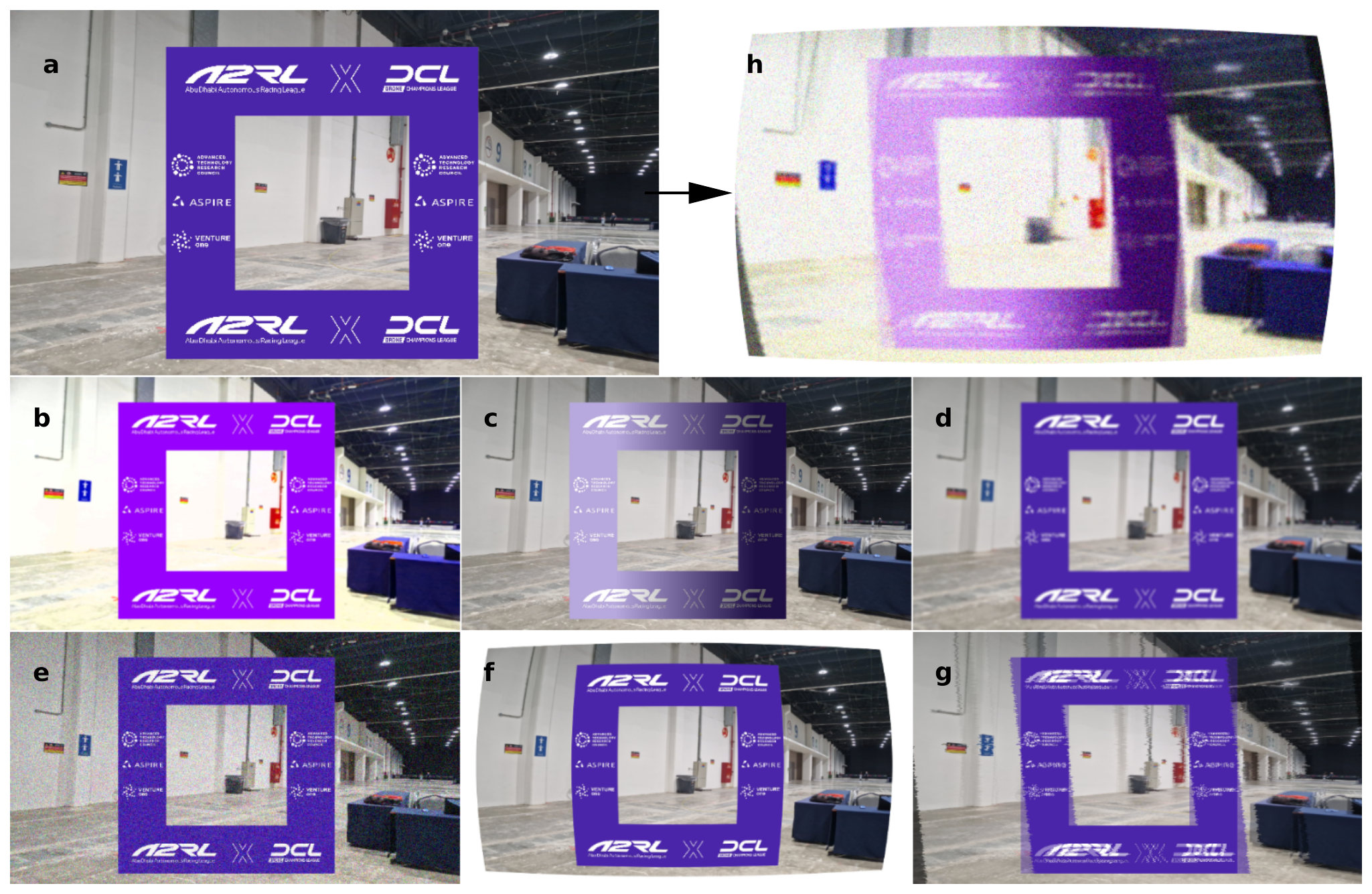}
  \end{minipage}

\raggedright

 \begin{minipage}[b][3.8cm][t]{0.02\textwidth}
    \textbf{C}
  \end{minipage}
  \begin{minipage}[b][4.0cm][t]{0.6\textwidth} 
\fbox{\footnotesize
\begin{tabular}{lcccc}
\multicolumn{1}{c}{\textbf{}}       & Resized & \begin{tabular}[c]{@{}c@{}}Resized\\ fixed AR\end{tabular} & \begin{tabular}[c]{@{}c@{}}Center\\ cropping\end{tabular} & \begin{tabular}[c]{@{}c@{}}Adaptive\\ cropping\end{tabular} \\
\hline
\multicolumn{1}{c}{IoU}             & 0.47    & 0.48                                                       & 0.36                                                      & \textbf{0.50}                                               \\
Detected corners                 & 19168   & 19436                                                      & 15437                                                     & \textbf{26516}                                              \\
Frames with PnP & 2950    & 2913                                                       & 2426                                                      & \textbf{3647}
\end{tabular}}
\end{minipage}
 \begin{minipage}[b][3.8cm][t]{0.02\textwidth}
    \textbf{D}
  \end{minipage}
  \begin{minipage}[b]{0.32\textwidth}
    \includegraphics[width=5.3cm, height=4.5cm]{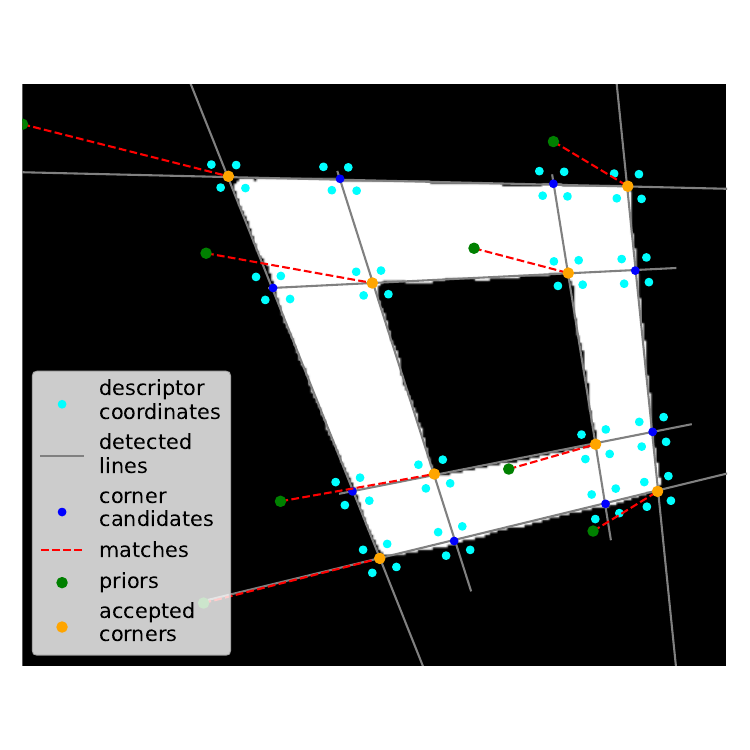}
  \end{minipage}

    \caption{(\textbf{A}) Vision Pipeline Elements. (\textbf{B}) Data augmentation from composited gate (a) and representative background. Individual augmentations include (b) HSV-based color augmentation, (c) directional brightness gradient, (d) Gaussian blur, (e) additive Gaussian noise, (f) lens distortion, and (g) rolling-shutter kernel blur. Combining these augmentations results in (h) fully augmented image.
    (\textbf{C}) Adaptive cropping allows light-weight processing of the most interesting part of the image, as illustrated by the best score in terms of IoU, number of accepted corners and `frames with PnP' computed for the three fastest successful flights.
    (\textbf{D}) `QuAdGate' precise corner fitting from possibly damaged segmentations based on edge detections. Lines intersections define precise corner candidates (blue). Around each candidate, local segmentation values are sampled (clockwise from the top-left) to form descriptors (cyan). These descriptors are then used to match candidates with the priors (green), yielding the final accepted corners (orange). Even with substantially misplaced priors (green), QuAdGate successfully matches them with the correct candidates (orange) while rejecting spurious corner candidates (blue).
    }
    \label{fig:vision}
\end{figure}

To obtain sub-pixel steady corners needed for precise PnP, despite sometimes rounded or incomplete gate segmentations, a solution is proposed that exploits the entire gate edge. 
The algorithm, called QuAdGate, extracts inner and outer gate corners from the segmentation masks (See Fig.~\ref{fig:vision}D). It detects and matches gate corners by combining the segmentation output with predicted corner coordinates (priors) based on state estimation. QuAdGate first uses a line detector to extract line segments from the segmentation mask and computes their intersections to identify corner candidates. It then matches these candidates to the priors using handcrafted descriptors obtained from the segmentation map.


\paragraph{Corner detection}
The segmentation mask is first derotated using the estimated attitude such that the vertical axis in the image matches the up-direction in the world frame. The corner detection begins by identifying lines using the OpenCV implementation of the Line Segment Detector (LSD) \cite{lsd_detector}, applied to the segmentation masks. In our implementation, the parameters are set to $\texttt{scale}=0.8$, $\texttt{sigma\_scale}=0.8$, $\texttt{quant}=25.0$, and $\texttt{ang\_th}=30.0$.
These are chosen to balance detecting the majority of lines from the segmentation masks while ensuring that each side of the gate is represented by exactly one line to avoid multiple detections. Next, the detected lines are extended equally in both directions such that their total length increases by a factor of $5/3$. The intersections between these extended lines are then computed to determine gate corner candidates. By relying on the intersections of the detected lines rather than the extremities of the segmentation mask itself, as in De Wagter et al. \cite{alphapilot_win_2019}, the accuracy of the corner coordinates is no longer affected by rounding or other minor imperfections in the mask.

\paragraph{Descriptors}
To associate corner candidates with prior gate corners projected from the EKF estimate,  handcrafted descriptors are extracted from the thresholded segmentation mask. Each descriptor captures four pixel values in the local neighborhood of a candidate corner:
\begin{align*}
d(p)=[v_{TL}, v_{TR}, v_{BR}, v_{BL}]^T
\end{align*}
where \(d(p)\) is the descriptor of corner candidate \(p\), and \(v_{\text{TL}}, v_{\text{TR}}, v_{\text{BR}}, v_{\text{BL}}\) are the corresponding pixel values at the top-left, top-right, bottom-right, and bottom-left, respectively. These positions are determined by moving $5$ pixels along the directions of the intersecting lines. For instance, the top-left value is obtained by moving 5 pixels to the left along the horizontal line and 5 pixels up along the vertical line. For a top-right outer corner, the resulting prior descriptor is $[0, 0, 0, 1]^T$, with $0$ representing regions of empty space and $1$ marking pixels belonging to the segmentation mask.

\paragraph{Corner matching}

An initial filtering step matches corner candidates to priors. Each candidate is first classified by corner type, allowing no discrepancy between the descriptor of a prior and that of a corner candidate. Candidates are further restricted to lie within $100$ pixels of the corresponding prior. Next, an optimal 2D affine transformation from priors to candidates, constrained to four degrees of freedom (translation, rotation, and uniform scaling), is estimated using RANSAC \cite{ransac} to filter out incorrect matches and other outliers. We employ OpenCV’s $\texttt{estimateAffinePartial2D}$ with $\texttt{RansacThreshold} = 5.0$. Lastly, the solution is rejected if the translation in any direction exceeds $150$ pixels, indicating poor convergence.

\subsubsection*{Perspective n Points}

In contrast to \cite{zurich_champion_level}, who solve a separate PnP problem for each gate, we combine the corners detected from multiple gates (two in our case) in a single PnP optimization when possible.
This approach offers two key advantages over treating each gate independently.

\begin{figure}[hbtp]
    \raggedright
 \begin{minipage}[b][5.0cm][t]{0.02\textwidth}
    \textbf{A}
  \end{minipage}%
  \begin{minipage}[t]{0.98\textwidth}
    \includegraphics[width=\textwidth]{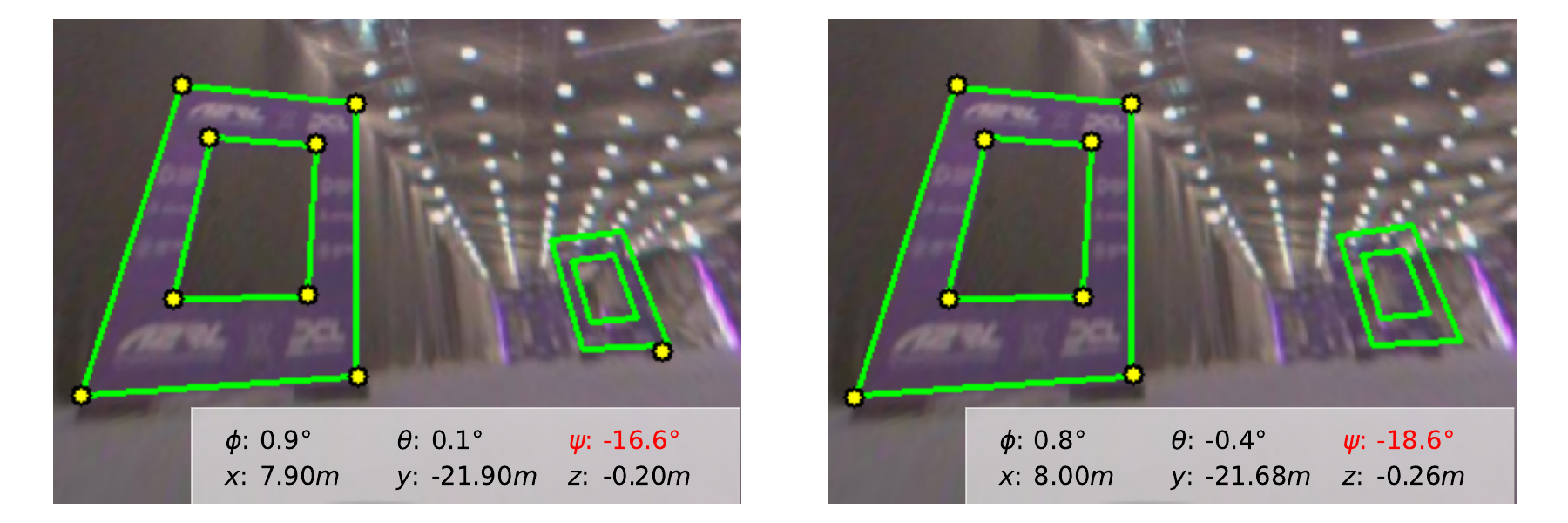}
  \end{minipage}

\raggedright
 \begin{minipage}[b][5.5cm][t]{0.02\textwidth}
    \textbf{B}
  \end{minipage}%
  \begin{minipage}[t]{0.98\textwidth}
    \includegraphics[width=\textwidth]{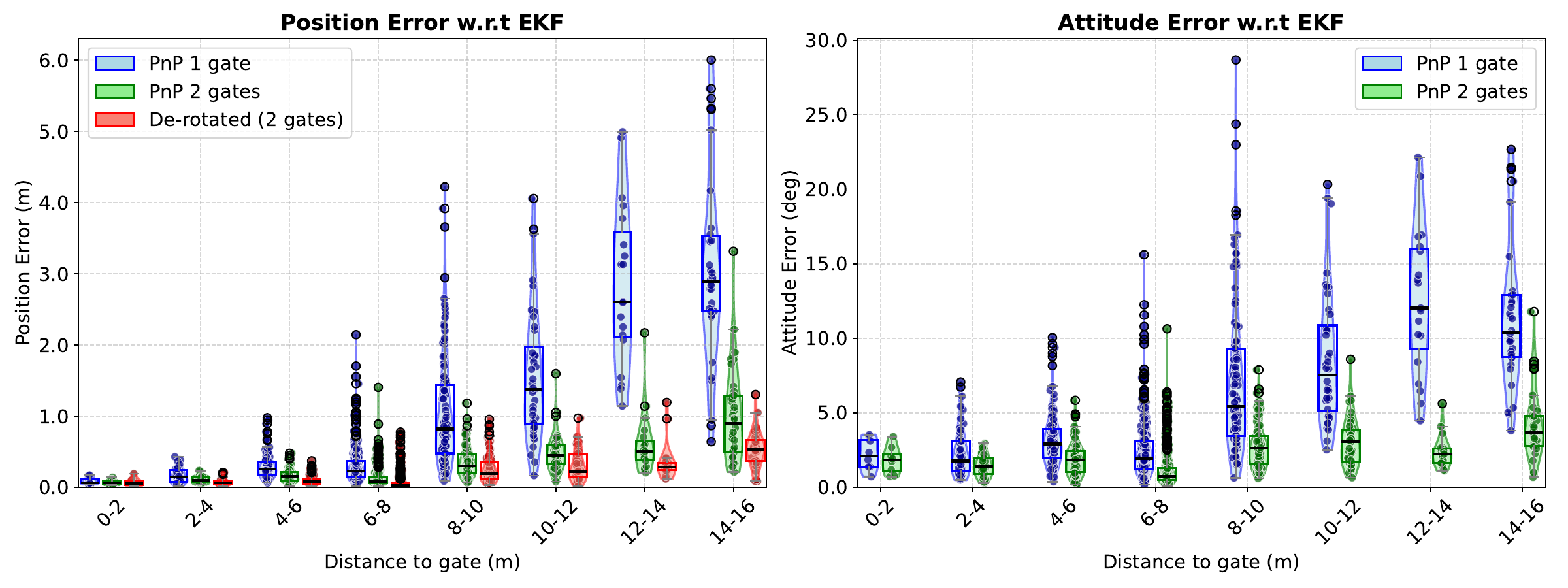}
  \end{minipage}

\raggedright
 \begin{minipage}[b][4.1cm][t]{0.02\textwidth}
    \textbf{C}
  \end{minipage}%
  \begin{minipage}[t]{0.98\textwidth}
    \centering
    \includegraphics[width=\textwidth]{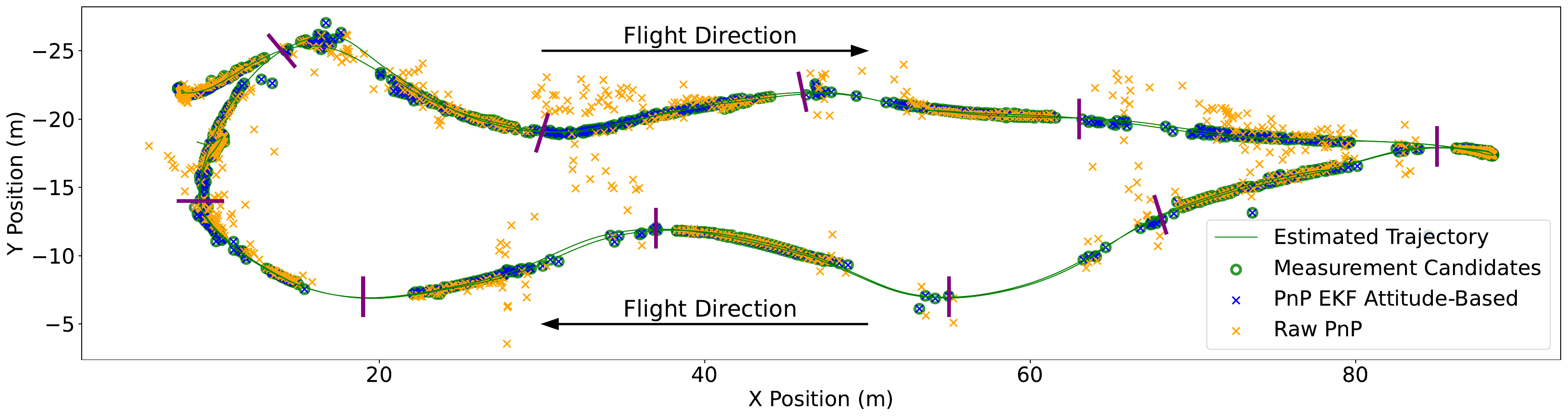}
  \end{minipage}

  \caption{Analysis of the PnP position and attitude accuracy in function of number of gates, gate distance and gate-derotation. (\textbf{A}) PnP heading initialization error depends on the number of gates. Even a single point on a second gate improves the reprojection. (\textbf{B}) PnP accuracy statistics for the three fastest successful. The accuracy becomes unacceptable for single gates beyond about 5m but remains good with derotated double gates. (\textbf{C}) Overview of the measurement candidates for the Kalman filter during the fastest successful flight, M16, against Killian Rousseau. When multiple gates are in view, the raw PnP becomes visibly better.}
  \label{fig:PnP_analysis}
\end{figure}

First, when the $2$D–$3$D correspondences come from gates located at different depths (non-coplanar), the variation in spatial configuration improves the ability of PnP to distinguish between translation and rotation. This is illustrated in Fig.~\ref{fig:PnP_analysis}a, where we show in green the reprojection of the first two gates using the estimated drone state based on the initial PnP with a different number of tracked corners. It can be seen that using even a single corner from a second gate improves this reprojection (which corresponds to a more accurate state) and results in a heading estimation difference of approximately $2^\circ$. This improvement is particularly important for long tracks, where initial heading errors strongly influence state estimation.

A second important advantage of merging corners from multiple gates in a single PnP optimization is that it allows PnP to reach its required threshold of at least four points more often. For the fastest successful flight, we detected enough corners for PnP $963$ out of $1620$ frames, of which $263$ ($27\%$) were multi-gate PnP cases. Among these, only $82$ instances had two gates with at least four detected corners each, while $181$ frames benefited from partial detections.
Sparse corner detections occur most frequently at distant gates, as the wide camera field of view and limited pixel resolution make segmentation and corner detection challenging.
In Fig.~\ref{fig:PnP_analysis}b we show that PnP nevertheless benefits from the partial gate corner detections in our three fastest flights. It can be seen that including these additional corners from different gates reduces both position and attitude errors. The difference is most visible when the distance to the primary gate is large.

The OpenCV implementation returns the camera position in a world coordinate system by combining the relative pose to the gate with the gate’s known world coordinates. The solution remains sensitive to geometry, especially at long viewing distances to gates as small rotational errors, for instance due to remaining delays combined with fast body rotations, can still result in important position errors.
Furthermore, when the gate is very close, motion blur degrades the quality of gate segmentation, resulting in less reliable corner detections. Lastly, PnP is less reliable when using a small number of points. To address this, we only use the full PnP solution when the gate distance is within a range of 2
m to 5 m and when a sufficient number of corners (at least six) are detected. If one of the two conditions is not met, we instead use the attitude from the Kalman filter state combined with the relative translation (w.r.t. the gate) estimate from PnP to compute the drone's world position. This fallback approach ensures a more accurate position estimation, especially for far gates, as shown in the left plot in Fig.~\ref{fig:PnP_analysis}b, denoted as "De-rotated (2 gates)". In Fig.~\ref{fig:PnP_analysis}c, we show the position estimate from OpenCV (Raw PnP) for each frame with a successful PnP and the de-rotated EKF Attitude method for our fastest flight. In green, we highlight the estimates that were used in our Kalman Filter as measurement candidates.  This fallback approach ensures a more stable and
robust position estimation. Especially far away from the next gate, the blue crosses much better represent the drone's position than the highly noisy orange crosses.

Since our position estimate is strongly coupled with our (EKF) attitude estimation, several measures are needed to feed minimal errors in the state estimation. Therefore, we only update our attitude from PnP in three scenarios that provide sufficiently reliable attitude updates:
\begin{enumerate}
    \item When the distance to the gate is between 2 m and 5 m, and more than six corners are detected.
    \item When at least two gates are detected, which improves the ability to distinguish between translation and rotation, making the PnP attitude estimate reliable enough for attitude updates (see right plot in Fig.~\ref{fig:PnP_analysis}b).
    \item When the drone is stationary (not in flight). In this case, we use the PnP attitude to correct for any potential misalignment in the initial pose estimate, particularly in the otherwise unobservable drone’s heading.
\end{enumerate}

\subsection*{State Estimation}

\subsubsection*{IMU Saturation}

During our high-speed flights, aggressive maneuvers reaching up to $7g$, combined with structural vibrations and variations in IMU quality across different drones, pushed the accelerometers beyond their $16g$ measurement range. When saturation occurs, the sensor output is not consistently limited to $\pm16g$, but often is inconsistent, making such events more difficult to detect. This saturation or corruption of accelerometer data caused divergence in the Kalman filter, almost inevitably leading to crashes.

To overcome these sensor limitations and enable higher-speed flight, we implemented an onboard algorithm that continuously compares predicted accelerations from a dynamic drone model with measured accelerations during all high-speed competition flights. The predicted accelerations are derived from the quadcopter model used in our RL simulator, as defined in Equation \ref{eq:thrust_drag_model}. When the difference between the predicted and measured accelerations exceeds a predefined threshold, the system switches to using the model-based acceleration predictions for state estimation in the Kalman filter.

To reduce high-frequency noise, the raw IMU accelerations are smoothed with a first-order low-pass (exponential moving-average) filter. The identical filter is applied to the model-predicted accelerations so that both signals experience the same time delay:

\begin{equation}
\mathbf{a}_{\text{filt}}[n] = 
    \alpha\,\mathbf{a}_{\text{filt}}[n\!-\!1] 
    \;+\; (1-\alpha)\,\mathbf{a}[n]
\end{equation}

For the IMU saturation detection, we compare the filtered sensor output with the filtered model estimate. If the Euclidean norm of their difference exceeds a heuristic threshold $\sigma=22\ \text{m/s}^2$, we switch to the model-based predicted accelerations (in all 3 axes) for our state prediction in the KF:

\begin{equation}
\bigl\lVert \mathbf{a}_{\text{filt}}^{\text{model}} - \mathbf{a}_{\text{filt}}^{\text{IMU}} \bigr\rVert_2
\;>\; \sigma .
\end{equation}

\subsubsection*{Kalman Filter}
We implement an Extended Kalman Filter (EKF) for fusing IMU measurements with position and attitude measurements obtained from the PnP solutions.

We define the state vector \(\mathbf{x} \in \mathbb{R}^{16}\) as:
\[
\mathbf{x} =
\begin{bmatrix}
x & y & z & v_x & v_y & v_z & q_w & q_x & q_y & q_z & b_x & b_y & b_z & b_p & b_q & b_r
\end{bmatrix}^\top
\]
The control input vector \(\mathbf{u} \in \mathbb{R}^{6}\) consists of IMU measurements. As described before, we use the modeled acceleration instead of the acceleration measured from the accelerometer in cases where the accelerometer saturates.

\[
\mathbf{u} =
\begin{bmatrix}
a_x & a_y & a_z & p & q & r
\end{bmatrix}^\top
\]
The process noise vector \(\mathbf{w} \in \mathbb{R}^{12}\) captures IMU noise and bias drift:
\[
\mathbf{w} =
\begin{bmatrix}
w_x & w_y & w_z & w_p & w_q & w_r & w_{bx} & w_{by} & w_{bz} & w_{bp} & w_{bq} & w_{br}
\end{bmatrix}^\top
\]
The continuous-time process model is given by:
\[
\dot{\mathbf{x}} = 
\mathbf{f}_c(\mathbf{x}, \mathbf{u}, \mathbf{w}) =
\begin{bmatrix}
    \dot{\mathbf{p}} \\
    \dot{\mathbf{v}} \\
    \dot{\mathbf{q}} \\
    \dot{\mathbf{b}}
\end{bmatrix} =
\begin{bmatrix}
\mathbf{v} \\
\mathbf{R}(\mathbf{q}) \begin{bmatrix}
    a_x - b_x - w_x &&
    a_y - b_y - w_y &&
    a_z - b_z - w_z
\end{bmatrix}^\top + \mathbf{g} \\
\frac{1}{2} \mathbf{q} \otimes
\begin{bmatrix}
    0 &&
    p-b_p-w_p &&
    q-b_q-w_q &&
    r-b_r-w_r
\end{bmatrix}^\top \\
\begin{bmatrix}
    w_{bx} & w_{by} & w_{bz} & w_{bp} & w_{bq} & w_{br}
\end{bmatrix}^\top
\end{bmatrix}
\]
where:
\begin{itemize}
    \item \(\mathbf{p} = [x, y, z]^\top\) is the position in world coordinates,
    \item \(\mathbf{v} = [v_x, v_y, v_z]^\top\) is the velocity in world coordinates,
    \item \(\mathbf{q} = [q_w, q_x, q_y, q_z]^\top\) is the orientation quaternion,
    \item \(\mathbf{b} = [b_x, b_y, b_z, b_p, b_q,b_r]^\top \) are the accelerometer and gyro biases,
    \item \(\mathbf{R}(\mathbf{q})\) is the rotation matrix from body to world frame,
    \item \(\mathbf{g} = [0, 0, g]^\top\) is the gravity vector,
    \item \(\otimes\) denotes quaternion multiplication.
\end{itemize}
We discretize the process model with forward euler to obtain the discrete state transition model:
\[
\bm{x}_{k+1} = \mathbf{f}(\bm{x}_k, \bm{u}_k, \bm{w}_k) =  \bm{x}_k + \mathbf{f}_c(\bm{x}_k, \bm{u}_k, \bm{w}_k) \cdot \Delta t_k
\]
The measurement model corresponds to position and orientation estimates obtained from the PnP algorithm:
\[
\mathbf{h}(\mathbf{x}) = 
\begin{bmatrix}
x & y & z & q_w & q_x & q_y & q_z
\end{bmatrix}^\top
\]
The EKF performs a prediction step for every incoming IMU measurement \(\mathbf{u}_k\):
\begin{align*}
\mathbf{x}_{k+1} &= \mathbf{f}(\mathbf{x}_k, \mathbf{u}_k, \mathbf{0}) \\
\mathbf{P}_{k+1} &= \mathbf{F}_k \mathbf{P}_k \mathbf{F}_k^\top + \mathbf{L}_k \mathbf{Q}_k \mathbf{L}_k^\top
\end{align*}

An update step is performed whenever a new PnP measurement \(\mathbf{z}_k = [x, y, z, q_w, q_x, q_y, q_z]^\top\) becomes available:
\begin{align*}
\mathbf{x}_{k+1} &= \mathbf{x}_{k} + \mathbf{K}_k \left( \mathbf{z}_k - \mathbf{h}(\mathbf{x}_k) \right) \\
\mathbf{P}_{k+1} &= \left( \mathbf{I} - \mathbf{K}_k \mathbf{H}_k \right) \mathbf{P}_{k}
\end{align*}
Here:
\begin{itemize}
    \item $\mathbf{K}_k$ is the Kalman gain calculated by $\mathbf{K}_k = \mathbf{P}_{k} \mathbf{H}_k^\top \left( \mathbf{H}_k \mathbf{P}_{k} \mathbf{H}_k^\top + \mathbf{R}_k \right)^{-1}$
    \item \( \mathbf{F}_k = \left. \frac{\partial \mathbf{f}}{\partial \mathbf{x}} \right|_{\mathbf{x}_k, \mathbf{w}_k=0} \) is the Jacobian of the process model w.r.t the state,
    \item \( \mathbf{L}_k = \left. \frac{\partial \mathbf{f}}{\partial \mathbf{w}} \right|_{\mathbf{x}_k, \mathbf{w}_k=0} \) is the Jacobian of the process model w.r.t. process noise,
    \item \( \mathbf{Q}_k \), \( \mathbf{R}_k \) are the process and measurement noise covariances,
    \item \( \mathbf{H}_k = \frac{\partial \mathbf{h}}{\partial \mathbf{x}} \) is the measurement Jacobian,
\end{itemize}

Accurate sensor fusion in our visual-inertial pipeline requires temporal synchronization between the camera and the IMU data. Our system establishes this by employing a timestamping mechanism and compensating for transport and processing latencies using pre-calibrated delays. Upon the arrival of each image, the correctly timestamped sensor measurements are fused to update a state estimate corresponding to the image's time of capture. To provide the flight controller with low-latency pose estimates, this delayed state is propagated forward using all subsequent IMU measurements that have not yet been fused. In Software-In-The-Loop (SITL) simulations on the A2RL competition track, with simulated delays of 17ms for images and 0.5ms for IMU data, our method reduced the RMS trajectory error to 0.103m. This presents a significant improvement over the 0.289m error from a baseline approach that naively processes the image-derived pose updates with the most recent IMU measurements at their time of arrival.

The measurement noise of the PnP estimation is modeled as:
\vspace{1em}

\begin{minipage}{0.45\textwidth}
\begin{equation}
\sigma_{\text{pos}}^2 = \frac{0.02 \cdot d_{\text{gate}}^2}{N_c^2 \cdot N_g}
\end{equation}
\end{minipage}
\hfill
\begin{minipage}{0.45\textwidth}
\begin{equation}
\sigma_{\text{quat}}^2 = \frac{0.01 \cdot d_{\text{gate}}^2}{N_c^2 \cdot N_g}
\end{equation}
\end{minipage}

\vspace{1em}

Outlier rejection is performed based on the Kalman filter position estimate, its covariance, and the number of detected corners. A PnP measurement is accepted if the following condition holds:

\begin{equation}
\left\| \mathbf{x}_{\text{pos}} - \mathbf{x}_{\text{PnP}} \right\|^2 
<  16 \cdot N_c^2  \cdot \text{trace}(\mathbf{P}_{\text{pos}})
\end{equation}

where $\mathbf{x}_{\text{pos}}$ is the position estimate from the Kalman filter, $\mathbf{x}_{\text{PnP}}$ is the position estimated from PnP, and $\mathbf{P}_{\text{pos}}$ is the position covariance matrix from the Kalman filter.

\subsection*{Control}

We use G\&CNets \cite{izzo2024optimality} to directly compute low-level motor commands from the predicted states provided by the vision pipeline. These networks are trained using Proximal Policy Optimization (PPO) in a simulation environment. The following sections detail our reinforcement learning (RL) pipeline, which includes a detailed model for the quadcopter (\textit{Quadcopter Model}), an initialization strategy designed to cover a wide range of flight conditions (\textit{Initialization}), parameter randomization to improve sim-to-real transfer (\textit{Domain Randomization}), the design of our reward function (\textit{Reward Function}), and the policy architecture and training specifics (\textit{Policy and Training}).

\subsubsection*{Quadcopter Model}
\label{sec:quadcopter_model}
To train the neural control policies, we use a simulation model of the quadcopter, similar to the one described in \cite{ferede2025one}. The state vector $\mathbf{x}$ and control input vector $\mathbf{u}$ of the quadcopter are defined as follows:
\[
\mathbf{x} =
\begin{bmatrix}
x & y & z & v_x & v_y & v_z & q_w & q_x & q_y & q_z & p & q & r & \omega_1 & \omega_2 & \omega_3 & \omega_4
\end{bmatrix}^\top
\]
\[
\mathbf{u} =
\begin{bmatrix}
u_1 & u_2 & u_3 & u_4
\end{bmatrix}^\top
\]
Here, $\bm{p} = (x, y, z)$ and $\bm{v} = (v_x, v_y, v_z)$ represent the position and linear velocity of the quadcopter in the world frame (North-East-Down, NED, convention), respectively. The unit quaternion $\bm{q}=(q_w, q_x, q_y, q_z)$ represents the orientation, while $\bm{\Omega} = (p, q, r)$ are the body angular rates. The variables $\bm{\omega} = (\omega_1, \omega_2, \omega_3, \omega_4)$ denote the angular velocities of the individual motors. The control input $\mathbf{u}$ consists of the motor commands applied to each rotor.

The equations of motion of the quadcopter are defined as:
\begin{align*}
\dot{\bm{p}} &= \bm{v} \\
\dot{\bm{v}} &= R(\bm{q}) \bm{F} + \bm{g} \\
\dot{\bm{q}} &= \frac{1}{2} \bm{q} \otimes 
\begin{bmatrix}
0 & p & q & r
\end{bmatrix}^\top \\
\dot{\Omega} &= \bm{M} \\
\dot{\omega} &= (
\bm{\omega_{c}} - \bm{\omega})/\tau
\end{align*}
Here $R(\bm{q})$ is the rotation matrix from body to world frame and $\bm{g} = [0, 0, g]^\top$ is the gravitational acceleration vector. $\bm{F} = [D_x, D_y, T]^\top$ contains the (specific) thrust and drag forces in the body frame. The models for these forces follow previous work \cite{ferede2024supervised, ferede2024end}, but include two extensions motivated by data collected at higher flight speeds: a quadratic drag term $v|v|$, and additional dependencies involving the advance ratio $\mu$ and the angle off attack $\alpha$ \cite{svacha2017improving}. These effects become increasingly relevant at the high speeds reached with more aggressive policies.

The resulting expressions for the specific forces are
\begin{equation}
    \bm{F} = \begin{bmatrix}
    -k_x v^B_x \sum_{i=1}^4 \omega_i - k_{x2} v^B_x |v^B_x| \\
    -k_y v^B_y \sum_{i=1}^4 \omega_i - k_{y2} v^B_y |v^B_y| \\
    -k_\omega \left(1 + k_{\alpha} \alpha + k_{\text{hor}} \mu \right) \sum_{i=1}^4 \omega_i^2
    \end{bmatrix}
    \label{eq:thrust_drag_model}
\end{equation}
Here, $\mathbf{v}^B$ denotes the velocity $\mathbf{v}$ expressed in the body frame. The angle of attack of the propeller blade is given by $\alpha$, and $\mu$ represents the effective advance ratio of the blade. These quantities are defined as follows:
\begin{align*}
    \alpha = \tan^{-1}\Big(\frac{v^B_z}{r \bar{\omega}}\Big) \quad \mu = \tan^{-1}\Big(\frac{v^{B2}_x+v^{B2}_y}{r \bar{\omega}}\Big) \quad \text{with} \quad \bar{\omega}=\sum_{i=1}^4 \omega_i
\end{align*}
where $r$ is the propeller radius\footnote{$r$ was estimated to be $0.0485775$ and is not randomized.}, $\omega_i$ are the individual motor angular velocities, and $\bar{\omega}$ is the total rotational speed across all four motors.

The angular acceleration $\bm{M} = (M_x$, $M_y$, $M_z)$ around the body axes are modeled as
\begin{equation}
    \bm{M} = \begin{bmatrix}
-k_{p1} \omega_1^2 - k_{p2} \omega_2^2 + k_{p3} \omega_3^2 + k_{p4} \omega_4^2 + J_x q r \\
-k_{q1} \omega_1^2 + k_{q2} \omega_2^2 - k_{q3} \omega_3^2 + k_{q4} \omega_4^2 + J_y p r \\
-k_{r1} \omega_1 + k_{r2} \omega_2 + k_{r3} \omega_3 - k_{r4} \omega_4 - k_{r5} \dot{\omega}_1 + k_{r6} \dot{\omega}_2 + k_{r7} \dot{\omega}_3 - k_{r8} \dot{\omega}_4 + J_z p q
    \end{bmatrix}
\label{eq:moment_model}
\end{equation}
Each moment component is expressed as a weighted combination of squared rotor speeds, capturing how differential thrust produces roll, pitch, and yaw moments. Gyroscopic coupling terms are also included. These terms become increasingly important at high rotational rates due to the drone’s elongated mass distribution—with the battery mounted at the front and the Jetson Orin at the rear—which introduces inertia asymmetry and makes the coupling terms non-negligible. The values of $J_x$, $J_y$, and $J_z$ were identified experimentally through a free-fall spinning throw of the drone without propellers. In this experiment, the gyroscope directly measures the cross-coupling effects, enabling estimation of the inertia-difference terms via linear regression.

Similar to \cite{ferede2025one} the steady-state motor response $\boldsymbol{\omega}_c$ is modeled as
\begin{align*}
    \omega_{c,i} = (\omega_{\text{max}} - \omega_{\text{min}})\sqrt{k u_i^2 + (1 - k_l) u_i} + \omega_{\text{min}}
\end{align*}

The nominal parameter values used in the quadcopter model are listed in Table~\ref{tab:policy_overview}.
In simulation, these equations of motion are discretized using forward Euler integration with a time step of $\Delta t = 0.01\,\text{s}$.

\subsubsection*{Initialization}
To ensure broad coverage of possible flight conditions, we use two initialization schemes: one that places the drone on the ground near the first gate, and one that samples uniformly from the full flight volume.

In the \textit{ground initialization}, the drone's position is sampled around a nominal starting point \((8, -22, 0)\) with uniform noise in the horizontal directions:
\[
x_0, y_0 \sim \mathcal{U}(-0.5, 0.5) + (8, -22), \quad z_0 = 0.
\]
The drone starts with zero linear and angular velocity:
\[
\bm{v}_0 = \bm{0}, \quad \phi_0 = \theta_0 = 0,
\]
\[
\psi_0 \sim \mathcal{U}\left(-\frac{\pi}{4}, \frac{\pi}{4}\right) + \psi_{\text{gate}}, \quad \bm{\Omega}_0 = \bm{0}, \quad \bm{u}_0 = \bm{0}
\]

In the \textit{uniform initialization}, the drone is randomly placed anywhere in the flight arena:
\[
x_0 \sim \mathcal{U}(1, 95), \quad y_0 \sim \mathcal{U}(-27, 1), \quad z_0 \sim \mathcal{U}(-5, 0),
\]
\[
\bm{v}_0 \sim \mathcal{U}(-0.5, 0.5)^3, \quad
\phi_0, \theta_0 \sim \mathcal{U}\left(-\frac{\pi}{9}, \frac{\pi}{9}\right), \quad
\psi_0 \sim \mathcal{U}(-\pi, \pi),
\]
\[
\bm{\Omega}_0 \sim \mathcal{U}(-0.1, 0.1)^3, \quad
\bm{u}_0 \sim \mathcal{U}(-1, 1)^4.
\]
The target gate is chosen as the nearest gate located ahead of the drone, i.e., with the drone positioned behind its plane.
\subsubsection*{Domain Randomization}
To improve sim-to-real transfer, we apply parameter randomization at the beginning of each episode. The nominal parameters of the A2RL quadcopter are identified using linear regression on high-speed flight data (see Table~\ref{tab:policy_overview}). Similar to the approach in~\cite{ferede2025one}, we randomize each parameter by a specified percentage around its nominal value. The level of randomization is selected based on a trade-off between robustness and performance. Table~\ref{tab:policy_overview} lists the ranges used for the competition policies. In all policies except M16, each parameter is sampled independently. For this exception, yaw moment model parameters are randomized jointly:
\[
    k_{r1} = k_{r2} = k_{r3} = k_{r4} \sim \mathcal{U}(\mathrm{min}, \mathrm{max}), \quad
    k_{r5} = k_{r6} = k_{r7} = k_{r8} \sim \mathcal{U}(\mathrm{min}, \mathrm{max}).
\]
The corresponding table entries for $k_{r2}$--$k_{r4}$ and $k_{r6}$--$k_{r8}$ are marked with '--' to indicate equality rather than independent sampling.

\subsubsection*{Reward Function}
The reward function balances task completion, flight smoothness, and perceptual robustness. All components are defined at each timestep \( k \). A detailed overview of the parameter values used in the expressions of the reward function can be found in Table \ref{tab:policy_overview}.

The \textit{progress reward} encourages movement toward the target gate:
\[
r_{\text{prog},k} = \lambda_{\text{prog}} \min\left( \left\| \mathbf{p}_{k-1} - \mathbf{p}_{g_k} \right\| - \left\| \mathbf{p}_{k} - \mathbf{p}_{g_k} \right\| , v_{\max} \Delta t\right)
\]
where \( \mathbf{p}_k \) is the drone’s position and \( \mathbf{p}_{g_k} \) the center of the current gate. This type of progress-based reward is widely used in reinforcement-learning approaches to drone racing \cite{Autonomous_Drone_Racing_with_Deep_Reinforcement_Learning, zurich_champion_level, ferede2024end, ferede2025one, OCvsRL}. To encourage safer behavior for slower policies, we cap the maximum achievable progress at $ v_{\max} \Delta t$. This constraint limits the effective maximum speed of the policy and discourages overly aggressive or unsafe trajectories.

The \textit{gate reward} provides a bonus upon successfully passing a gate:
\[
r_{\text{gate},k} = \lambda_{\text{gate}} \quad \text{if gate is passed, otherwise } 0.
\]

The \textit{angular rate penalty} discourages high angular velocities:
\[
p_{\text{rate},k} = \lambda_{\text{rate}} \left\| \boldsymbol{\Omega}_k \right\|^2.
\]

The \textit{gate offset penalty} discourages off-center gate crossings:
\[
p_{\text{offset},k} = \lambda_{\text{offset}} \left\| \mathbf{p}_k - \mathbf{p}_{g_k} \right\| \quad \text{if gate is passed, otherwise } 0.
\]

The \textit{perception penalty} encourages the drone to keep the next gate within the forward-facing camera’s view:
\[
p_{\text{perc},k} = \lambda_{\text{perc}} \, \theta_{\text{cam}} \quad \text{if } \theta_{\text{cam}}> \pi/3, \text{ otherwise } 0,
\]
where \( \varphi \) is the angle between the optical axis and the center of the next gate.

The \textit{motor penalty} discourages large changes in motor commands:
\[
p_{\Delta \bm{u},k} = \lambda_{\Delta \bm{u}} \sum_{i=0}^{4} \max\left( \left| u_i[k] - u_i[k-1] \right| - \Delta u_{\text{thresh}}, 0 \right).
\]
This penalty was introduced after observing that an early policy exhibited highly aggressive bang--bang control, switching almost exclusively between minimum and maximum motor commands, which led to motor overheating.
The \textit{low action penalty} discourages low actions:
\[
p_{\bm{u},k} = \lambda_{\bm{u}} \sum_{i=0}^{4} \max (0.5-u_i, 0).
\]
This penalty was initially added to further reduce bang--bang behavior and was used for one of the networks used in competition. However, in practice, the \textit{motor penalty} proved more effective, and it became our primary method for regularizing the control inputs.
The \textit{crash penalty} applies a fixed penalty when a collision occurs:
\[
p_{\text{crash},k} = \lambda_{\text{crash}} \quad \text{if collision, otherwise } 0.
\]

The total reward at timestep \( k \) is:
\[
r_k = r_{\text{prog},k} + r_{\text{gate},k} - p_{\text{rate},k} - p_{\text{offset},k} - p_{\text{perc},k} - p_{\Delta \bm{u},k} - p_{\bm{u},k} - p_{\text{crash},k}.
\]

\subsubsection*{Collision detection and gate passing}
In simulation, the drone is modeled as a point mass. The gate’s inner size varies during training and is denoted by $g_{\mathrm{size}}$, while the outer size is fixed at $2.7\,\mathrm{m}$ and the gate thickness at $g_{\mathrm{thickness}}$. All gates share the same inner size, except during the training of policy M16, where gates 2 and 6 were reduced to $0.45\,\mathrm{m}$. Additionally, the gate thickness was reduced to $0.8,\mathrm{m}$ for policy M16.

A gate collision is registered when the drone intersects the gate’s collision box or crosses the gate plane outside the inner square opening (i.e., more than $g_{\mathrm{size}}/2$ from the center along either axis). A ground collision occurs when $z < -h_{\mathrm{ground}}$ while the drone’s speed exceeds $v_{\mathrm{ground}}$.  An out-of-bounds event occurs when $x \notin [1,95]$ or $y \notin [-27,1]$, or when the angular velocity exceeds $1700\,\mathrm{deg/s}$.

In the reinforcement learning setup, an episode is terminated whenever a gate collision or an out-of-bounds event occurs.

\subsubsection*{Policy and Training}
The selected neural policy is a three-layer fully connected network with ReLU activation functions and 64 neurons per layer (See Fig~\ref{fig:overview}). The policy takes in 24 observations, including the quadcopter's state and information about current and future gates, similar to prior work~\cite{ferede2025one}:
\[
\bm{x}_{\text{obs}} = [\bm{p}^{g_i}, \bm{v}^{g_i}, \bm{\Phi}^{g_i}, \bm{\Omega}, \bm{\omega}, \bm{p}_{g_{i+1}}^{g_i}, \psi_{g_{i+1}}^{g_i}]^T
\]
Here, $g_i$ denotes the reference frame of the $i$-th gate. The vector $\bm{p}^{g_i}$ is the position in the current gate frame, $\bm{v}^{g_i}$ the velocity, $\bm{\Phi} = (\phi, \theta, \psi)$ the Euler angles, $\bm{\Omega}$ the angular velocity in the world frame, and $\bm{\omega}$ the angular velocity in the body frame. The components $\bm{p}_{g_{i+1}}^{g_i}$ and $\psi_{g_{i+1}}^{g_i}$ represent the relative position and yaw angle of the next gate, expressed in the current gate frame.

Although the simulator uses quaternions internally, we represent orientation using Euler angles in the observation vector. The network outputs four motor commands $\bm{u}$.

The quadcopter model, initialization scheme, domain randomization, and reward function described above are used to construct a custom Gym environment for training. We train the policy using the Proximal Policy Optimization (PPO) algorithm~\cite{ppo}, implemented via the Stable-Baselines3 library~\cite{stable-baselines3}. Table~\ref{tab:policy_overview} provides an overview of the reward function parameters, collision detection settings, PPO training parameters, and domain randomization used for the competition policies.


\clearpage
\bibliography{science_template}
\bibliographystyle{sciencemag}


\section*{Acknowledgments}

We gratefully acknowledge the many participants of previous races whose dedication and efforts laid the foundation for this work. Their contributions created the knowledge base and inspiration that made further progress possible. We also sincerely thank ADR and AIRR for their help in pushing this field of research. Finally, we would like to thank A2RL and DCL for this opportunity.

\paragraph*{Funding:}

Parts of this work, in particular, the development of the RL-based control, were performed with the support of `SPEAR'  project nr 101119774 under HORIZON-CL4-2022-DIGITAL-EMERGING-02-06.





\end{document}